\documentclass[11pt]{article}

\usepackage[T1]{fontenc}
\usepackage[utf8]{inputenc}
\usepackage{lmodern}
\usepackage[margin=0.82in]{geometry}
\usepackage{microtype}
\usepackage{graphicx}
\usepackage{booktabs}
\usepackage{tabularx}
\usepackage{longtable}
\usepackage{array}
\usepackage{amsmath,amssymb}
\usepackage{enumitem}
\usepackage{xcolor}
\usepackage{hyperref}
\usepackage{url}
\usepackage{float}
\usepackage{caption}
\usepackage{natbib}
\usepackage{fancyhdr}
\definecolor{praxaNavy}{HTML}{17324D}
\definecolor{praxaBlue}{HTML}{2F6B8A}
\definecolor{praxaTeal}{HTML}{2A7F73}
\definecolor{praxaGold}{HTML}{B7791F}
\definecolor{praxaRed}{HTML}{A23B3B}
\definecolor{praxaGray}{HTML}{65717C}

\hypersetup{
  colorlinks=true,
  linkcolor=praxaBlue,
  citecolor=praxaTeal,
  urlcolor=praxaBlue,
  pdfauthor={Stefan G. Creadore},
  pdftitle={From Proposal to Verified Effect: Praxa, an Evidence-Bound Harness for Governed AI Agent Execution},
  pdfsubject={AI agent execution governance, engineering validation, and benchmark methodology},
  pdfkeywords={AI agents, agent harness, governance, verification, provenance, evaluation}
}

\setlist{nosep,leftmargin=*}
\newcolumntype{Y}{>{\raggedright\arraybackslash}X}
\newcommand{\pass}{\textcolor{praxaTeal}{\textbf{PASS}}}
\newcommand{\hold}{\textcolor{praxaRed}{\textbf{HOLD}}}
\newcommand{\notest}{\textcolor{praxaGold}{\textbf{NOT TESTED}}}
\newcommand{\repo}[1]{\texttt{#1}}

\title{\textbf{\textcolor{praxaNavy}{From Proposal to Verified Effect: Praxa, an Evidence-Bound Harness for Governed AI Agent Execution}}\\[0.45em]
\large An Engineering Validation, Descriptive-Pilot, and Benchmark-Governance Study}
\author{\textbf{Stefan G. Creadore}\\Independent Researcher, Valrico, Florida, United States\\
\href{mailto:stefan@nex-t1.ai}{stefan@nex-t1.ai}\quad\textbullet\quad
\href{https://orcid.org/0000-0003-2268-053X}{ORCID 0000-0003-2268-053X}\\
\href{https://praxa.io}{praxa.io}\quad\textbullet\quad
\href{https://docs.praxa.io}{docs.praxa.io}\quad\textbullet\quad
\href{https://github.com/praxa-labs/praxa-benchmarks}{Public artifacts}}
\date{July 22, 2026; revised July 27, 2026\\[0.4em]\small Preprint --- intended for arXiv submission (cs.AI / cs.SE)}

\begin{document}
\maketitle

\begin{abstract}
Large-language-model agents can propose and execute actions, but proposal, authority, dispatch, verified external effect, and serving promotion are different claims. We present Praxa, an agent harness that represents these states explicitly through deterministic admission, brokered execution, external read-back, reconciliation, and reviewed promotion. We report four evidence lanes. First, an author-run repository-local audit at a pinned revision passed 1,027/1,027 unit tests and 89/89 Workerd tests, instrumented all 363 expected source files, and met four coverage floors; raw per-test transcripts and independent reproduction are unavailable. Second, in a provider-backed Terminal-Bench Core 0.1.1 pilot across 12 curated tasks, baseline and reliability-layer arms each passed 17/36 strict trials. The reliability layer used 37.49\% more input and 50.73\% more output tokens, so the pilot does not support superiority. Third, in a post-debug, two-order coordination-proxy development comparison, baseline and a source-authored candidate each completed 180/180 trials with equal measured accuracy, full hermetic crash recovery, and zero protected violations. The candidate used 37.11\% fewer tokens, 33.84\% lower estimated endpoint cost, and 11.63\% fewer steps; this does not establish improved quality, latency, or production behavior. Fourth, deployed source/configuration evidence shows bounded reflection, recall accounting, memory compilation, and tool-health paths, but no production outcome lift. Praxa's supported contribution is an evidence-bound architecture that makes authority-to-effect transitions explicit and testable. Current evidence does not establish adversarial security, production safety, general specialist superiority, autonomous recursive optimization, or user benefit.
\end{abstract}


\section{Introduction}

Language models become operational agents when a surrounding system gives model outputs consequences. A model may choose a tool, formulate arguments, delegate work, or recommend a transaction; the harness determines whether those proposals become effects. This distinction matters wherever agents touch communication, commerce, personal data, browsers, enterprise records, or long-running workflows. Fluent text can conceal the difference among an intention, an authorized request, an acknowledged dispatch, and a verified outcome.

Prior work has shown the value of interleaving reasoning and action \citep{yao2023react}, learning when to invoke tools \citep{schick2023toolformer}, adding reflective feedback \citep{shinn2023reflexion}, and programming multi-agent conversations \citep{wu2024autogen,hong2024metagpt}. Security systems now provide deterministic privilege control, capability and information-flow isolation, and least-privilege tool scopes \citep{shi2025progent,debenedetti2025camel,zhu2025miniscope}. These are close and important antecedents. Praxa does not claim to invent deterministic authorization. It studies the adjoining systems problem: after admission, how should dispatch, uncertain outcomes, external read-back, reconciliation, evidence retention, and serving promotion remain distinguishable? Multi-agent failure analyses further show that orchestration and verification defects are often system-design failures rather than model-only failures \citep{cemri2025failures}.

Praxa addresses this engineering boundary. Its public integration surface consists of an SDK, command-line interface, MCP-facing contracts, OpenAPI operations, and durable mission abstractions. An access-controlled authority plane retains deterministic policy, provider credentials, broker execution, evidence adapters, and rollout authority. The intended contract is concise:

\begin{center}
\small\textbf{model or client proposes $\rightarrow$ policy authorizes $\rightarrow$ broker executes}\\
\textbf{$\rightarrow$ evidence verifies $\rightarrow$ review promotes.}\normalsize
\end{center}

The architecture draws on fail-safe defaults, complete mediation, and separation of privilege \citep{saltzer1975protection}; per-request authorization in zero-trust systems \citep{rose2020zerotrust}; provenance as an explicit relation among entities, activities, and agents \citep{w3c2013prov}; and lifecycle-oriented risk governance \citep{tabassi2023airmf}. We call the resulting sequence the \emph{authority--effect chain}. Its research value is the explicit refusal to collapse six state claims into one: proposed, admitted, dispatched, externally verified, reconciled, and promoted.

The central methodological problem is that agent-system claims frequently collapse distinct questions:

\begin{enumerate}
  \item Does source code encode the intended rule?
  \item Do deterministic tests exercise that rule?
  \item Does it hold at the target runtime boundary?
  \item Does a controlled model-provider experiment show task benefit?
  \item Has an independently reviewed candidate been authorized to serve?
  \item Does deployed observation show real-world benefit or harm reduction?
\end{enumerate}

A test pass cannot answer the fifth or sixth question. A benchmark score cannot mint execution authority. A deployment receipt cannot establish model quality. Praxa therefore treats source conformance, runtime validation, provider evaluation, promotion, and production observation as separate evidence classes.

This paper makes seven bounded contributions:

\begin{enumerate}
  \item an evidence-bound reference architecture separating proposal, admission, dispatch, external-effect verification, uncertainty, reconciliation, and serving authority;
  \item an executable authority model based on canonical actions, scope intersection, approvals, idempotency, budgets, revocation, and fail-closed transitions;
  \item a pinned release-validation snapshot with test counts, coverage numerators and denominators, source-inventory completeness, and cryptographic evidence identifiers;
  \item a controlled Terminal-Bench pilot reporting an observed aggregate tie, corrected pass@$k$ semantics, missing-scorer sensitivity, and token cost without inferential claims;
  \item a post-debug provider-backed coordination-proxy development comparison reporting matched conformance and bounded resource reductions without inferring model-quality, latency, production, or recursive-optimizer superiority;
  \item a source-bound audit of deployed learning-loop wiring plus a development-only hybrid tool-candidate-generation evaluation with explicit statistical insufficiency and a separately reported runtime-union result; and
  \item frozen confirmatory protocols in which adversarial authority/effect trials, full-suite reliability evaluation, synthetic conformance, provider qualification, and production promotion remain distinct evidence classes.
\end{enumerate}

\section{Research questions and hypotheses}

\subsection{Research questions}

\textbf{RQ1.} Can an agent harness represent proposal, admission, dispatch, externally verified effect, unresolved uncertainty, reconciliation, and serving authority as distinct, fail-closed contracts?

\textbf{RQ2.} Does the audited Praxa snapshot satisfy its declared repository-local validation contract?

\textbf{RQ3.} Does a reliability layer consisting of budget-aware planning, verification, checkpoint/rollback, and recovery improve task completion under a controlled terminal-agent pilot?

\textbf{RQ4.} Can a bounded recursive harness protocol aggregate fixed-context evidence and select a candidate while preserving an explicit production-ineligible state?

\textbf{E1 (post hoc exploratory question).} In a post-debug provider-backed development comparison, does the source-authored coordination candidate preserve observed conformance while reducing measured resource use?

\begin{table}[H]
\centering
\caption{Hypotheses and their disposition, revised July 27, 2026. ``Supported'' is bounded to the named evidence class; E1 is post hoc and is not added as a hypothesis.}
\label{tab:hypotheses}
\small
\begin{tabular}{p{0.07\textwidth} p{0.58\textwidth} p{0.25\textwidth}}
\toprule
\textbf{ID} & \textbf{Hypothesis} & \textbf{Disposition} \\
\midrule
H1 & Proposal and execution authority remain structurally separate across gateway, policy, broker, verifier, reconciliation, and rollout paths. & Supported by author-run repository-local source/test mapping; not a security proof. \\
H2 & The pinned snapshot satisfies its defined regression and inventory contract. & Supported by the author-run aggregate for the audited revision. \\
H3 & The reliability layer improves Terminal-Bench trial accuracy under the pilot design. & Not supported: 17/36 in both arms. \\
H4 & Synthetic recursive conformance can exercise pairing and selection without conferring production eligibility. & \pass{} for deterministic gate mechanics only. \\
\bottomrule
\end{tabular}
\end{table}

Three confirmatory questions remain open: whether the authority--effect chain prevents adversarial unauthorized effects without unacceptable utility loss; whether a preregistered full-suite reliability treatment improves task completion or efficiency; and whether deployed Praxa improves operational or user outcomes. These are protocol targets, not evaluated hypotheses. The repository also contains preliminary TLA+ design modules, but no checked model-checking result was available. We therefore make no formal-verification claim.

\section{Related work}

\subsection{Reasoning, acting, tools, and multi-agent systems}

ReAct interleaves natural-language reasoning with actions in external environments \citep{yao2023react}. Toolformer studies models that learn when and how to invoke APIs \citep{schick2023toolformer}. Reflexion stores linguistic feedback to alter later attempts without updating model weights \citep{shinn2023reflexion}. AutoGen and MetaGPT make role structure and interaction topology programmable components of multi-agent applications \citep{wu2024autogen,hong2024metagpt}. Automated design of agentic systems goes further by searching over agent programs \citep{hu2025adas}; Recursive Harness Self-Improvement revises task-specific prompt-level harnesses using feedback over revision history \citep{lee2026rhi}. Praxa adopts the premise that harness structure is consequential, but treats an optimizer's output as an untrusted candidate: it cannot modify tools, secrets, budgets, approval policy, tenant scope, or rollout authority.

\subsection{Agent evaluation}

AgentBench evaluates agents across interactive environments \citep{liu2024agentbench}, while SWE-agent demonstrates that the agent-computer interface materially affects software-engineering performance \citep{yang2024sweagent}. $\tau$-bench introduces tool-agent-user interaction under domain policies and emphasizes repeated-trial reliability \citep{yao2025taubench}. Terminal-Bench uses containerized terminal tasks and execution-based tests \citep{terminalbench2025core,merrill2026terminalbench}. HELM argues for multi-metric, transparently configured evaluation rather than single-score reporting \citep{liang2023helm}. AutoDojo shows that adaptive black-box attacks can expose weaknesses missed by static prompt-injection evaluation \citep{ma2026autodojo}. These works motivate fixed task and model bindings, separate effectiveness and resource metrics, explicit invalid-outcome handling, and the frozen protocol's sealed adversarial trials. A genuinely adaptive attacker with a bounded query budget remains necessary future work; it is not part of the present protocol.

Praxa's repository-local test counts are not agent capability scores. Conversely, the Terminal-Bench pilot is a provider experiment but does not validate Praxa's production deployment. Keeping these lanes separate is part of the paper's method.

\subsection{Risk, adversarial environments, and human oversight}

ToolEmu uses an LM-emulated sandbox to scale risk discovery \citep{ruan2024toolemu}; AgentDojo tests agents against prompt injection in tool-returned data \citep{debenedetti2024agentdojo}. Progent supplies a policy language and deterministic privilege mediation for tool calls \citep{shi2025progent}. CaMeL extracts trusted control and data flows and applies capabilities so untrusted data cannot redirect protected flows within its threat model \citep{debenedetti2025camel}. MiniScope reconstructs permission hierarchies and computes least-privilege authorization for tool-calling agents \citep{zhu2025miniscope}. AI-control work studies monitoring and intervention when a capable model may be intentionally subversive \citep{greenblatt2024control}. Human--AI interaction guidelines emphasize making capabilities, uncertainty, correction, and user control legible \citep{amershi2019guidelines}.

Table~\ref{tab:closest-work} states the intended distinction conservatively. Praxa is broader in lifecycle span but currently weaker in adversarial evidence than these security-focused systems. In particular, this paper reports neither CaMeL-style guarantees nor a Progent- or MiniScope-scale security/utility comparison.

\begin{table}[H]
\centering
\caption{Positioning against the closest authorization and prompt-injection systems. ``Not primary'' means the cited work does not make that stage its principal evaluated contribution, not that an implementation cannot support it.}
\label{tab:closest-work}
\scriptsize
\setlength{\tabcolsep}{3pt}
\begin{tabular}{p{0.12\textwidth} p{0.23\textwidth} p{0.18\textwidth} p{0.20\textwidth} p{0.17\textwidth}}
\toprule
\textbf{System} & \textbf{Primary boundary} & \textbf{Deterministic control} & \textbf{Uncertain effect / read-back} & \textbf{Evidence-governed promotion} \\
\midrule
Progent & Tool and argument privileges & Policy-language mediation & Not primary & Not primary \\
CaMeL & Trusted control/data flow and capabilities & System-enforced flow isolation & Not primary & Not primary \\
MiniScope & Least-privilege tool permissions & Reconstructed scopes and authorization & User confirmation; effect lifecycle not primary & Not primary \\
Praxa & Proposal through verified effect and serving state & Canonical admission and broker boundary & Explicit uncertain state, verifier, and reconciliation & Separate reviewed transition; current evidence remains pre-production \\
\bottomrule
\end{tabular}
\end{table}

\subsection{Distributed effects, durable workflows, and artifact promotion}

The effect half of Praxa has an equally mature systems lineage. Sagas decompose long-lived transactions into committed subtransactions and compensating actions \citep{garciamolina1987sagas}. Work on applications beyond global distributed transactions emphasizes stable identities, explicit message/state boundaries, and application-level recovery \citep{helland2007life}. Beldi demonstrates fault-tolerant and transactional composition for stateful serverless workflows \citep{zhang2020beldi}. Likewise, signed evidence binding is established software-supply-chain practice: in-toto records cryptographically verifiable links across independently operated build and delivery steps \citep{torresarias2019intoto}. Praxa does not claim to invent compensation, idempotency, durable execution, reconciliation, provenance, or signed promotion. Its narrower contribution is their composition with an LLM-specific authority boundary and a claim discipline in which model proposal, external effect, evaluation evidence, and serving authority cannot substitute for one another. The current study validates pieces of that composition; it does not establish novel distributed-systems guarantees.

\subsection{Governance and provenance}

NIST AI RMF separates governance, mapping, measurement, and management activities \citep{tabassi2023airmf}. W3C PROV supplies a general vocabulary for provenance \citep{w3c2013prov}. Praxa operationalizes a narrower claim: every authority-bearing transition should be attributable to an authenticated principal, canonical action, deterministic policy decision, broker result, and evidence record. Hashes provide artifact identity and tamper evidence; they are not third-party certification.

\section{Praxa architecture and threat model}

\subsection{Actors and trust boundaries}

The public abstraction distinguishes six logical actors: (i) an authenticated principal; (ii) a client, model, or specialist that proposes work; (iii) a gateway that derives tenant, principal, purpose, and delegated scope; (iv) deterministic policy and budget logic; (v) a private broker that owns provider credentials; and (vi) an evidence path that checks external or system-visible state. A separate rollout authority determines which reviewed candidate may serve.

The client, model, optimizer, tool descriptions, and external content are untrusted with respect to authority. The gateway, policy owners, credential broker, verifier credentials, and signing controls are distinct trusted boundaries. The phrase ``independent verification'' in this paper means independent from the proposal's assertion and authority path; it does not always imply a different foundation model. In the Terminal-Bench pilot, actor and verifier used the same recorded provider alias with zero recorded fallbacks, but an immutable provider-side snapshot was not archived; the final task outcome was determined separately by benchmark tests.

\begin{figure}[H]
\centering
\includegraphics[width=\textwidth]{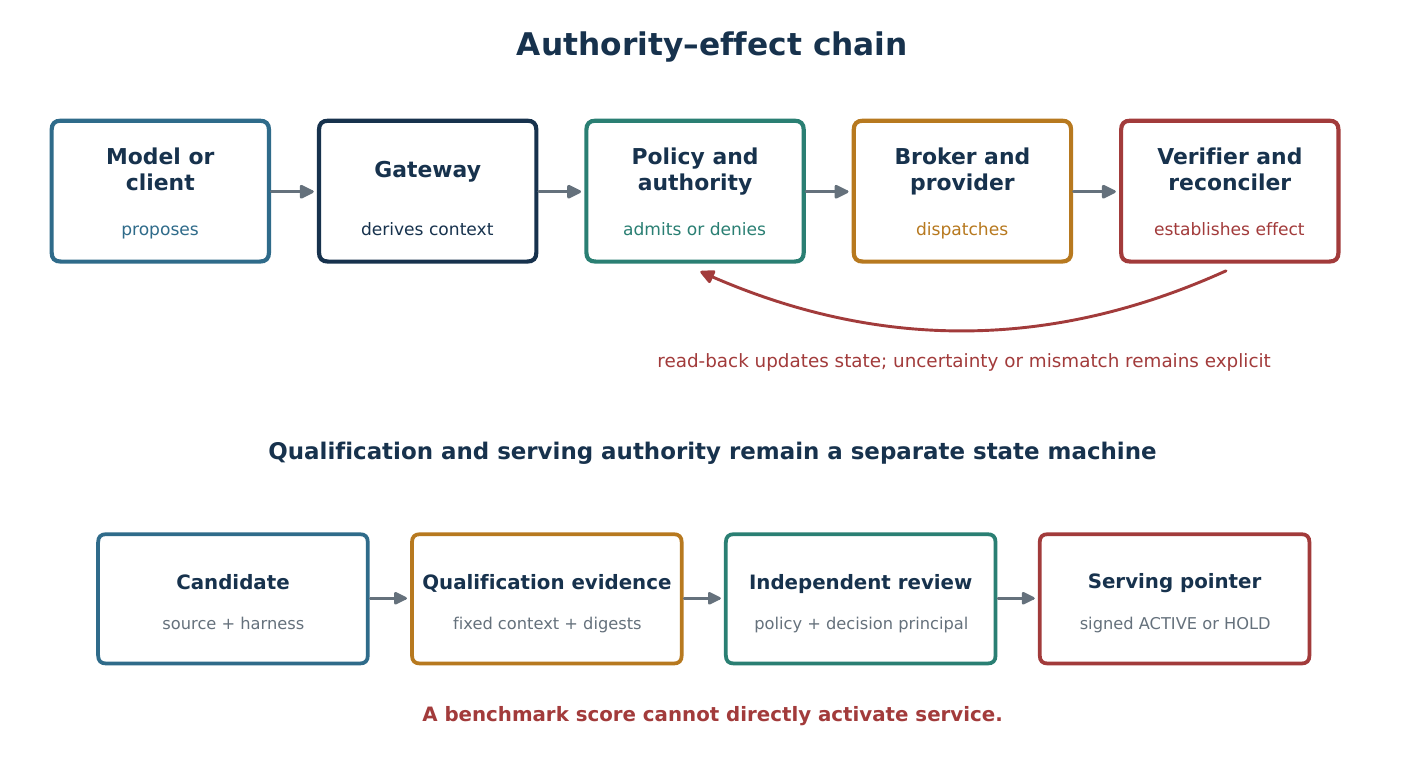}
\caption{Public abstraction of Praxa's authority--effect and qualification boundaries. The return path carries read-back evidence or uncertainty; it does not allow a model assertion to certify its own external effect. The lower path requires qualification evidence and independent review before a signed serving state. Internal service names, credentials, and exact deployment topology are intentionally outside the public artifact.}
\label{fig:architecture}
\end{figure}

\subsection{Proposal, authority, and effect states}

Let $p$ be a proposed canonical action and $c$ the server-derived context. Admission is modeled as a deterministic conjunction:

\begin{equation}
A(p,c)=I(c)\land T(c)\land S(p,c)\land C(p,c)\land B(p,c)\land R(c)\land F(p,c),
\end{equation}

where $I$ denotes authenticated identity, $T$ tenant/principal binding, $S$ delegated scope, $C$ consent or approval, $B$ available budget, $R$ freshness and non-revocation, and $F$ action-family and target policy. A failed term denies admission; a prompt cannot override it.

For a specialist or workflow, effective tools are an intersection:

\begin{equation}
\mathcal{T}_{\mathrm{eff}}=
\mathcal{T}_{\mathrm{runtime}}\cap
\mathcal{T}_{\mathrm{specialist}}\cap
\mathcal{T}_{\mathrm{skill}}\cap
\mathcal{T}_{\mathrm{user}}\cap
\mathcal{T}_{\mathrm{tenant}}\cap
\mathcal{T}_{\mathrm{workflow}}\cap
\mathcal{T}_{\mathrm{environment}}
\setminus \mathcal{D},
\end{equation}

where $\mathcal{D}$ is the union of explicit denials. No later layer may reintroduce a denied tool.

The effect lifecycle distinguishes:

\begin{equation}
\mathrm{proposed}\rightarrow
\left\{
\begin{array}{ll}
\mathrm{denied}, & A(p,c)=0,\\
\mathrm{admitted}\rightarrow\mathrm{dispatched}\rightarrow
\{\mathrm{verified},\mathrm{failed},\mathrm{uncertain}\}, & A(p,c)=1.
\end{array}
\right.
\end{equation}

An accepted request or provider acknowledgment is not automatically a verified effect. A timeout after dispatch remains \emph{uncertain} until a verifier can reconcile the intended action against provider-visible or system-visible state:

\begin{equation}
R(p,e)\in\{\mathrm{verified},\mathrm{failed},\mathrm{still\ uncertain}\},
\end{equation}

where $e$ is evidence obtained through a read path that does not merely repeat the proposer or dispatch response. Rewriting uncertainty as success would erase exactly the state that reconciliation needs. Mutations use idempotency identities, and approval-sensitive actions bind approval to a canonical action rather than a later paraphrase.

Serving authority is a second state machine, not the terminal state of the task evaluator:

\begin{equation}
\mathrm{candidate}\rightarrow\mathrm{qualified}\rightarrow\mathrm{reviewed}\rightarrow\mathrm{active}.
\end{equation}

Each transition binds an exact source revision, harness definition, evaluation context, evidence digest, and decision principal. A benchmark result may support qualification; it cannot directly produce \emph{active}. The implementation study evaluates five corresponding invariants: no proposal self-authorizes; no scope layer widens an earlier denial; no acknowledgment certifies its own external effect; unresolved uncertainty remains visible; and no synthetic or reporting-only score directly activates a candidate. These are tested engineering properties, not a formal proof that every implementation or trusted component is secure.

\subsection{Capabilities and invariants}

Table~\ref{tab:capabilities} summarizes the public-safe capability map. It describes architecture, not production availability.

\begin{table}[H]
\centering
\caption{Praxa harness capabilities and governing invariants.}
\label{tab:capabilities}
\small
\begin{tabular}{p{0.22\textwidth} p{0.34\textwidth} p{0.34\textwidth}}
\toprule
\textbf{Capability} & \textbf{Function} & \textbf{Invariant} \\
\midrule
Durable missions & Bounded, resumable work with cancellation and events & Mission acceptance is distinct from verified provider effect. \\
Tool and action registry & Canonical action identities, aliases, typed adapters & Aliases cannot widen canonical authority. \\
Policy and budgets & Identity, scope, approval, rate, cost, and target checks & Deterministic policy, not model prose, authorizes. \\
Specialists & Restricted roles, instructions, handoffs, and memory posture & A specialist is a restriction over existing execution owners. \\
Bounded learning loops & Typed negative-outcome reflection, recall accounting, concept compilation, and outcome-biased tool ranking & Deployed wiring is not evidence of correct lessons, improved recall, task benefit, or user benefit. \\
Brokered execution & Credential-isolated provider calls & Public packages and models do not hold provider authority. \\
Verification and reconciliation & Read-back or system-visible evidence & Record, rendered approval, and gate-match payload remain aligned. \\
Replay and provenance & Causal events, evidence digests, idempotency & Replay reconstructs recorded transitions; it does not recreate a stochastic world. \\
Rollout governance & Exact candidate, evidence, review, and serving pointer & Benchmark acceptance cannot directly activate service. \\
\bottomrule
\end{tabular}
\end{table}

\subsection{Public and access-controlled surfaces}

The public SDK, CLI, MCP contract library, and remote-gateway examples are maintained at \href{https://github.com/praxa-labs/praxa}{github.com/praxa-labs/praxa}. The public package describes governed MCP contracts but does not implement the private action executor. Product documentation is at \href{https://docs.praxa.io}{docs.praxa.io}, and the project site is \href{https://praxa.io}{praxa.io}. The access-controlled authority and execution implementation, which requires repository authorization, is referenced at \href{https://github.com/praxa-labs/praxa-harness}{github.com/praxa-labs/praxa-harness}. The publication artifacts and reviewed aggregates are at \href{https://github.com/praxa-labs/praxa-benchmarks}{github.com/praxa-labs/praxa-benchmarks} \citep{praxalabs2026artifacts,praxalabs2026docs}.

The public artifact excludes credentials, customer data, proprietary prompts, raw authority material, private fixtures, and unredacted provider transcripts. This boundary limits independent reproduction of private authority paths but avoids treating publication as permission to expose operational secrets.

\subsection{Multi-runtime utilization case study}

The broader Praxa AI application uses one agent identity across three intentionally different execution surfaces: cloud text, on-device Apple Foundation Models, and realtime voice. Capability parity is not assumed. Cloud text can expose a broad server-mediated catalog, on-device execution is limited by platform and privacy capabilities, and voice excludes visual or interaction-heavy tools that do not fit a hands-free runtime. This is a utilization example of the scope-intersection rule: a shared identity does not imply universal tool availability.

\begin{figure}[H]
\centering
\includegraphics[width=0.9\textwidth]{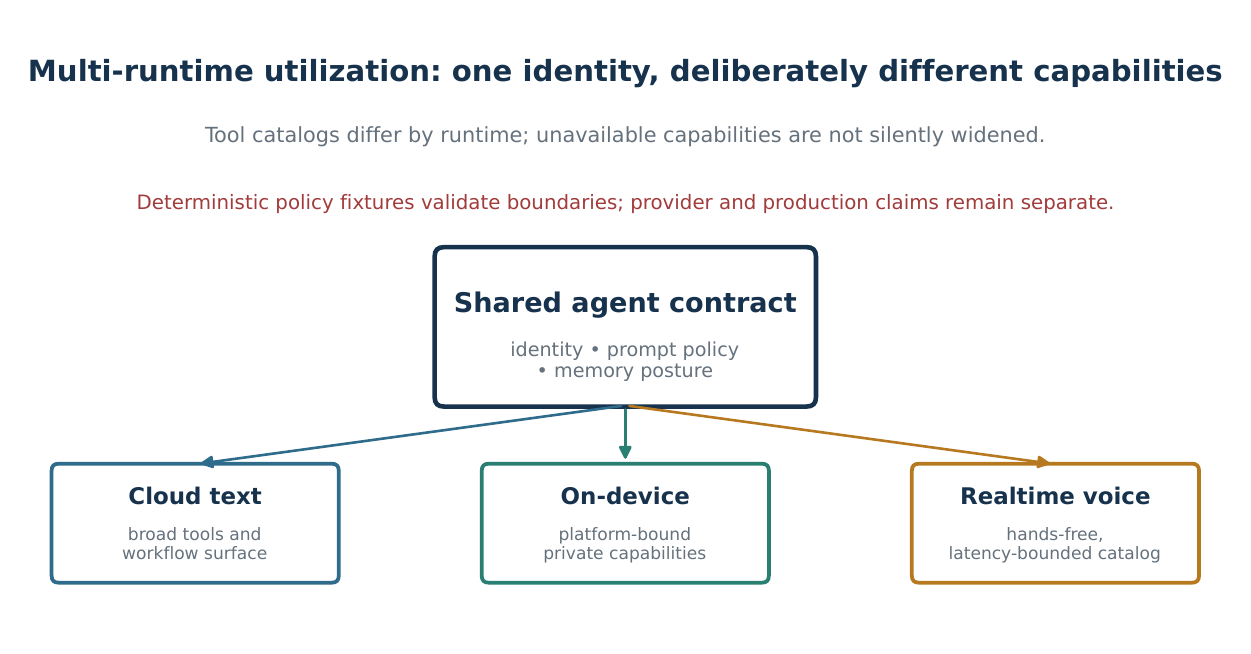}
\caption{Praxa AI utilization across cloud text, on-device, and realtime voice. The shared contract coordinates identity and policy; each runtime retains a deliberately restricted capability catalog.}
\label{fig:multi-runtime}
\end{figure}

The application also exercises durable research workflows, approval-gated browser actions, prompt routing, tool scoping, and specialist policy. A July 22 audit of that shared application tree observed the bounded evaluation inventory in Table~\ref{tab:utilization-evals}. The audited tree was dirty at revision \texttt{af6c494d}; the public machine-readable record contains the full hash. Its own artifact schema consequently classified the specialist source evaluation as non-publishable. We therefore include this inventory only as secondary utilization evidence, not as the primary reproducible result.

\begin{table}[H]
\centering
\caption{Secondary deterministic evaluation inventory from the multi-runtime application. Units have different oracles, overlap in code paths, and must not be summed.}
\label{tab:utilization-evals}
\scriptsize
\begin{tabular}{p{0.19\textwidth} p{0.16\textwidth} p{0.25\textwidth} p{0.30\textwidth}}
\toprule
\textbf{Evaluation} & \textbf{Observed unit} & \textbf{Measured property} & \textbf{Claim ceiling} \\
\midrule
Injection policy & 80 fixtures; 108 focused tests passed & Known attacks across seven surfaces and six vector families & Tested policy behavior; not a live penetration study. \\
Agentic routing & 34/34 deterministic cases & Routing, tool scope, and budget logic & No hosted-model quality result. \\
Prompt contracts & 61 contract items & Structure, placeholders, and target coverage & Not 61 model-response evaluations. \\
Affective uncertainty & 10/10 clean fixtures & Abstention and uncertainty labeling; zero unsupported hidden-emotion claims & No clinical or population validity. \\
Puppet browser replay & 9/9 scripted tasks over five fixtures & Deterministic browser-script replay & Not a hosted-model web benchmark. \\
Research workflow & 11 suites / 48 tests passed & Workflow and delivery contracts & No production answer-quality or delivery-success result. \\
Specialist policy & 16 manifests / 481 fixtures & Network-disabled authority and output-posture rules & No provider quality or deployment proof. \\
Hybrid tool selection & 114 inspected development queries / 133 tools & Candidate-stage compression plus separately measured full runtime-scope union & Development-only; labels were inspected during tuning and lack independent adjudication. \\
\bottomrule
\end{tabular}
\end{table}

This inventory illustrates why heterogeneous evaluation units should remain separate. A prompt-contract item, an injection fixture, a workflow test, and a provider-scored task do not share a denominator or scientific interpretation. At deployed revision \texttt{705abdce}, four bounded feedback paths were source/configuration-observed: negative-outcome reflection, memory-recall accounting, typed memory compilation, and tool-health ranking. This establishes available wiring, not invocation completion, a correct learned lesson, improved retrieval, task lift, or user benefit. The hybrid tool-selection evaluation distinguishes the bounded embedding-plus-intent candidate generator from the downstream deterministic runtime tool-scope union. It is also distinct from the separate fail-open Cohere/Workers-AI cross-encoder runtime; no isolated Cohere quality comparison or real-corpus memory-retrieval A/B is reported.

\section{Evidence model}

\begin{figure}[H]
\centering
\includegraphics[width=0.96\textwidth]{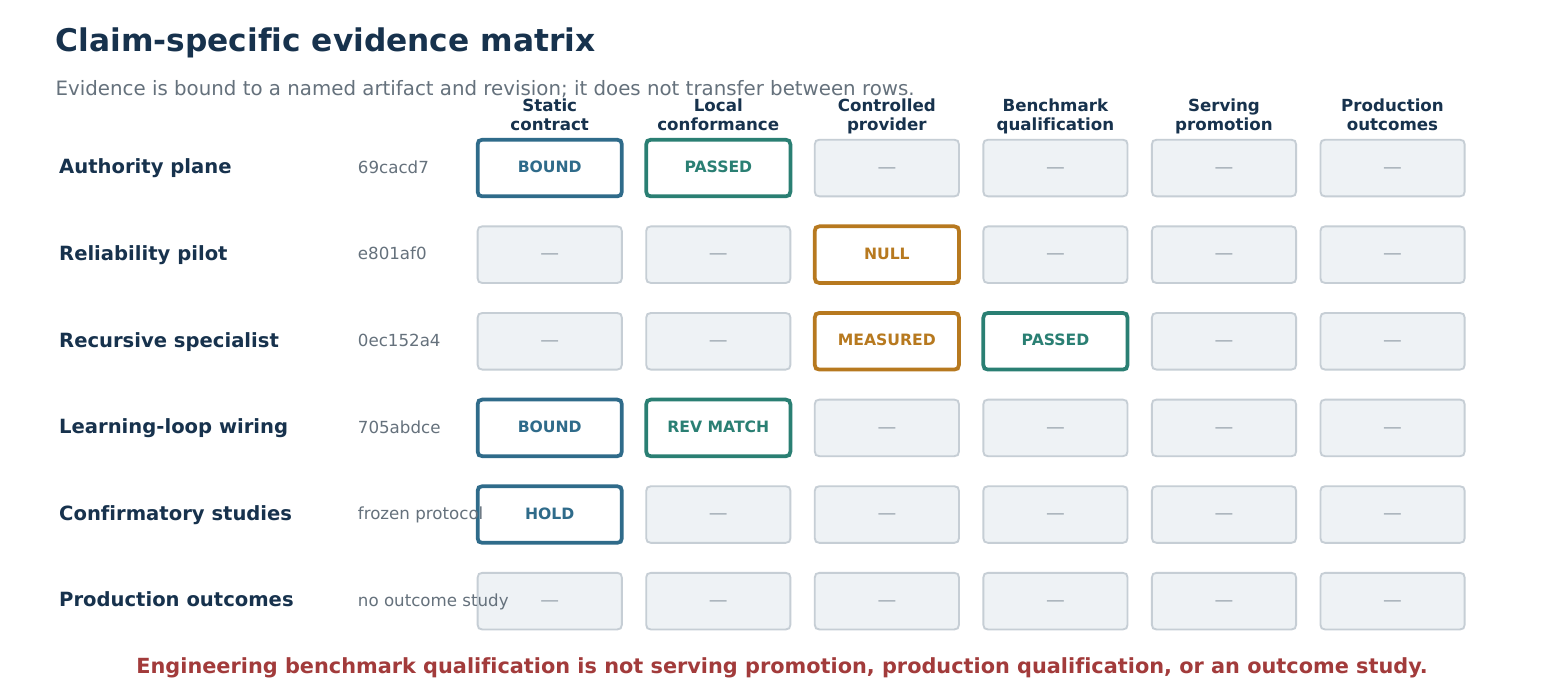}
\caption{Claim-by-evidence matrix. Each mark is attached to a named artifact and revision; marks are not cumulative across rows. The recursive specialist lane reaches a two-order provider-backed development comparison and counterbalance qualification at benchmark revision \texttt{0ec152a4}. Benchmark qualification is not serving promotion, production qualification, or production observation.}
\label{fig:ladder}
\end{figure}

Praxa uses claim-specific evidence lanes rather than a single project-wide maturity score (Figure~\ref{fig:ladder}). Within a lane, evidence may progress from a static contract to local conformance, a controlled experiment, benchmark qualification, separately authorized serving promotion, and production observation. Evidence does not transfer across lanes: a provider experiment on an earlier reliability harness cannot establish conformance of the current authority plane, repository-local tests cannot establish provider or production behavior, and an engineering benchmark gate cannot activate service.

\begin{enumerate}
  \item \textbf{Static contract:} types, schemas, inventories, and deterministic policy definitions.
  \item \textbf{Local conformance:} executable tests, runtime-boundary tests, coverage, and artifact binding at a source revision.
  \item \textbf{Controlled provider experiment:} fixed tasks, model binding, budgets, timestamps, outcome scorer, and raw-result hashes.
  \item \textbf{Benchmark qualification:} fresh immutable evidence recomputed against fixed effectiveness, protected-dimension, and resource gates.
  \item \textbf{Serving promotion:} independent review, signed transition authority, controlled rollout, and rollback evidence.
  \item \textbf{Production observation:} deployed health, reconciliation, incident, user, and outcome evidence.
\end{enumerate}

This lane-specific model prevents two category errors: more tests do not become a larger model benchmark, and a benchmark on one revision does not become evidence for another revision or production authorization.

\section{Methods}

\subsection{Study A: pinned repository-local conformance}

The audited harness checkout was clean at commit:

\begin{center}
\small\nolinkurl{69cacd7f5c01cec311aec05efa8e66edc7a117ca}
\end{center}

The tracked release record classified itself as a \repo{repository\_local\_regression\_baseline}. An author-run \repo{npm run validate:all} execution on July 22, 2026 reproduced the unit, Workerd, and coverage aggregates. The bound identifiers were:

\begin{itemize}
  \item evidence bundle: \small\nolinkurl{b204c62cc2b77f1027a01e2efd8c36ba1883a21f199379f81f654ce6ac1a676f}\normalsize;
  \item source inventory SHA-256: \small\nolinkurl{bf5ece7b6372fe12abdfffc8c279ff686e9e52220943eda7c953c9fbdf3c2b9e}\normalsize;
  \item test evidence SHA-256: \small\nolinkurl{875965d931945da425becaedd4aa290426d32769ec0d64229cd0c26f08886ff8}\normalsize.
\end{itemize}

The unit lane exercised policy, contract, registry, parsing, budget, authorization, evidence, and orchestration behavior. The Workerd lane executed boundary tests under the Cloudflare Workers-compatible runtime. Results are reported separately because a Node-oriented unit environment cannot establish Workerd behavior, and neither lane establishes a live deployment.

V8 coverage was aggregated over a declared release-critical source population. We report both numerator and denominator. A separate inventory assertion required all expected source files to appear in instrumentation output. Therefore, 363/363 means coverage-population completeness, not 100\% line coverage.

The broader validation pipeline also checked source-size policy, action registry relationships, additive migration inventory, repository-declared Worker/Queue/HOLD topology, documentation links, package contents, and dependency audit state. The released invariant map links five authority--effect claims to pinned source files, named test files, and intended oracle titles. The raw Vitest and Workerd reports were gitignored and are absent from the pinned tree, so those mappings are source-addressable author-run evidence rather than independently auditable per-assertion transcripts. These checks are useful for release conformance but are not field-performance metrics.

\subsection{Study B: Terminal-Bench reliability-layer pilot}

The provider-backed pilot compared two arms built on one ReAct-style solver:

\begin{itemize}
  \item \textbf{Baseline:} reliability policies disabled.
  \item \textbf{Reliability layer:} budget-aware planning, a self-verification loop, checkpoint/rollback around state changes, and failure recovery enabled.
\end{itemize}

Both arms used Terminal-Bench Core 0.1.1, the recorded model binding \nolinkurl{vercel_ai_gateway/openai/gpt-4.1-mini}, the same base prompt, containerized task execution, and the same configured limits of 40 steps, 420 seconds, and a 260,000-token soft loop budget. The actor/planner loop gated entry using accumulated charged tokens, but an admitted call could overshoot and verifier calls were not fully charged to the guard at this revision; this was not a strict per-call ceiling. Recorded provider-token totals exceeded 260,000 in one baseline and two reliability-layer trials. The curated subset contained 12 tasks; each arm ran three attempts per task, yielding 36 trials per arm. Temperature was 0.0, three trials ran concurrently within an arm, and the arms ran sequentially in fixed order (baseline, then reliability layer) without interleaving or randomization. The baseline run recorded \repo{no\_rebuild:false}; the later reliability-layer run recorded \repo{no\_rebuild:true}. This build/cache asymmetry was not experimentally isolated and prevents attributing all between-arm runtime differences solely to the reliability policies. The run window was 00:00:27--00:42:48 UTC on July 2, 2026. Terminal-Bench task tests, rather than the actor model's self-assessment, determined the final pass/fail result.

The primary descriptive outcome was strict end-to-end trial accuracy:

\begin{equation}
\widehat{a}=\frac{\sum_{t=1}^{12}\sum_{j=1}^{3}\mathbf{1}[\mathrm{pass}_{t,j}]}{36}.
\end{equation}

Let $c_t$ be the number of passes among $n=3$ attempts for task $t$. We recomputed pass@$k$ from task-level outcomes using the standard combinatorial estimator \citep{chen2021codex}

\begin{equation}
\widehat{\mathrm{pass@}k}=\frac{1}{12}\sum_{t=1}^{12}\left[1-\frac{\binom{3-c_t}{k}}{\binom{3}{k}}\right],\quad k\in\{1,2,3\},
\end{equation}

with an impossible combination contributing zero. For $k=1$, this equals trial accuracy; for $k=3$, it is the proportion of tasks solved at least once. The earlier rendered summary called the solved-at-least-once value simply \repo{pass\_at\_k}, while raw run metadata explicitly stored pass@2. The values were not rival computations of the same $k$; the rendered label had discarded $k$. The public aggregate now preserves pass@1, pass@2, and pass@3 separately.

Three attempts per arm for \path{cron-broken-network} had \repo{parser\_results:null}. The surrounding metadata identifies them as parser failures but does not recover an official pass/fail score. Our primary strict analysis conservatively treats them as non-passes. A post hoc complete-case diagnostic excludes only those records from the trial-rate denominator; it is not a replacement headline. We also report input and output tokens, rollback count, and verification rounds. We do not report an inferential test: the study used only 12 curated tasks, attempts are clustered within task, no minimum detectable effect was preregistered, and three attempts per task are insufficient for a robust population claim.

The raw-result hashes are:

\begin{itemize}
  \item baseline: \small\nolinkurl{b937e91b489b7359aeda1c6b2d694320a256077156a50e6911877148beb3ed3f}\normalsize;
  \item reliability layer: \small\nolinkurl{c4595d98fcd9e9485467c8b2278e4abc7f4d05c42498983714570819d1c78ddf}\normalsize.
\end{itemize}

The redacted aggregate records the exact source and aggregation revisions. Raw transcripts are not included in the public artifact.

\subsection{Study C: recursive specialist benchmark governance}

The recursive lane permits bounded changes to role grouping, role instructions, typed handoffs, and coordination hops. It cannot change tools, credentials, budgets, approval policy, tenant boundaries, model/source bindings for a comparison, or rollout authority. Its checked-in conformance artifact pairs baseline and candidate trials under fixed identifiers and recomputes completion, accuracy, eligible recovery, latency, tokens, cost, steps, and protected violations.

The synthetic artifact remains explicitly \repo{productionEligible:false}; its authored values validate selection mechanics and are not performance measurements. A later protected provider producer recorded five executions during iterative development. Two early executions reached provider evaluation but retained no public result archive. The first preserved execution retained baseline after a protected-veto decision, and the first redesigned execution did not satisfy the absolute-quality floor. After evaluation-affecting engineering changes, the fifth execution passed the engineering benchmark gate. This sequence makes the retained comparison post-debug development evidence rather than an untouched confirmatory evaluation.

The strongest retained comparison binds benchmark revision \texttt{0ec152a4}, 30 fixed fixtures, three repetitions, and two opposite endpoint orders: baseline-first and candidate-first. Combining the order artifacts yields 180 trials per endpoint: 150 deterministic-preflight and 30 model-assisted trials. The trace records 192 provider calls. The evaluated topology is classified as a \repo{synthetic\_coordination\_proxy}; the recovery oracle injected a hermetic crash after commit, and latency is a two-order counterbalanced shadow proxy. The production \repo{spawn\_agents} specialist topology and web/mobile serving surfaces were not executed. The candidate topology was source-authored; the configured optimizer was not invoked.

The baseline-first workflow completed successfully. The candidate-first workflow completed its provider execution and retained a checksum-bound artifact, but the workflow itself ended red when the then-current inline comparator reported \repo{harness\_counterbalance\_identity\_mismatch}. A later credential-free workflow at verification revision \texttt{393a4646} repaired that comparison path, checked both retained artifact sets, and produced the qualified counterbalanced aggregate. Reporting the original red workflow status alongside the later successful verification prevents retrospective relabeling of the execution record.

The public gate contract requires matched context; accepted candidate decisions; zero safety, contract, grounding, structured-output, authorization, and duplicate-effect violations; completion, mean-accuracy, and eligible-recovery rates of 1.0; absolute ceilings of 60,000/120,000 ms for p50/p95 latency, 5,000,000 tokens, \$100 estimated endpoint cost, and 10,000 steps; relative candidate ceilings of 1.1 for latency, tokens, and steps and 110\% for estimated cost; strict reduction in token, cost, and step summaries; and at least one strict raw-metric improvement in both held-out and adversarial partitions. Three evaluator models voted 2:1 for the candidate, with no tie or veto, while the deterministic protected and resource predicates remained authoritative. The decision is producer-recorded; this release recomputes the published aggregate predicates and binds the producer and verifier source hashes, but private completions prevent public re-adjudication of every raw trial.

Endpoint cost is an estimate, not a provider billing receipt. The historical runs bound the execution alias \repo{gpt-5.6-sol} to operator-supplied rates of \$6.25 per million input tokens and \$30 per million output tokens and summed per-event micro-dollar estimates; the released aggregate is not retroactively repriced. Across both orders, endpoint estimates were \$1.578215 for baseline and \$1.044142 for candidate. Total trace use, including evaluation, was 306,330 tokens and an estimated \$2.676842. The verifier could grant counterbalance qualification only: production qualification and serving-path qualification remained false and serving authority remained \repo{none}. The public redacted aggregate preserves workflow outcomes, metrics, model bindings, pricing provenance, qualification state, and retained-artifact hashes without publishing private completions.

\subsection{Study D: secondary learning-loop and tool-selection audit}

A source/configuration audit at deployed revision \texttt{705abdce} mapped four bounded feedback paths: negative-outcome reflection, recall accounting, typed memory compilation, and outcome-biased tool ranking. The audit asks whether the paths and enabling configuration exist at one deployed source revision; it defines no task, user, recall, or tool-success outcome estimand.

The associated hybrid tool-selection artifact covers a fixed 133-tool catalog and 114 inspected development queries, partitioned into 70 calibration and 44 inspected holdout cases. It distinguishes two stages. The bounded candidate generator combines embedding rank with deterministic query-aware intent; the downstream runtime then unions that set with deterministic tool-scope admission. The artifact reports both stages rather than treating candidate precision as end-to-end precision. Because cases and labels were inspected during tuning and lack independent adjudication, the artifact is marked \repo{confirmatory:false} and \repo{status:insufficient\_evidence}. The separate fail-open cross-encoder runtime is described as implemented functionality; no isolated reranker-quality or real-corpus memory-retrieval result is inferred.

\subsection{Prospective confirmatory studies (frozen; not executed)}

Two versioned protocols convert the principal evidence gaps into falsifiable studies. Both are prospective and remain \hold{}: zero confirmatory provider trials, zero promotion decisions, and zero production observations have been recorded under them.

The governance protocol compares an unmediated direct adapter in the model-free runtime stage (prompt-only direct execution in the separately activated provider stage), static gateway/RBAC, deterministic admission with brokered execution, and full Praxa against a stateful provider simulator. Its unit of inference is a sealed scenario cluster. Fourteen attack and fault families cover forged authority, scope and alias confusion, approval substitution, revocation races, duplicate delivery, ambiguous dispatch, false acknowledgments, malicious evidence, and promotion-receipt tampering. The protected prevention endpoint is scenario-level unauthorized external effect, established from append-only mutation, protected-read, disclosure, and egress ledgers rather than an arm's own message. Deterministic-admission arms G2 and G3 are each tested against G0 and G1; G3 must also be non-inferior to G2 within a proposed +1.5-percentage-point unauthorized-dispatch risk-difference margin. A co-primary effect-integrity endpoint requires G3 to reduce false success-equivalent effect claims relative to G2. A second co-primary endpoint requires oracle-observed legitimate effect completion in G3 to be non-inferior to G2 within a proposed five-percentage-point margin; an arm's own \repo{verified} label does not enter this utility endpoint. Both margins require an independent ex ante acceptable-risk/utility rationale before activation. All components must pass. Six mechanism ablations isolate canonicalization, approval binding, dispatch-time revocation, idempotency, independent read-back, and signed promotion.

The reliability protocol replaces the curated pilot with the official Terminal-Bench 2.1 population: all 89 tasks in the corrected release, five attempts per task and arm, unchanged official timeouts and resources, and 890 valid trials for one frozen candidate against baseline \citep{terminalbench2026v21}. Four pilot identifiers also occur in Terminal-Bench 2.1 and are excluded from development selection; the remaining eight require a pre-activation content-similarity audit, with any material overlap blocking activation until a disjoint development suite is bound. No Terminal-Bench 2.1 outcome may influence candidate selection. A development-only ACTIVE manifest binds D0--D7 without pretending that the selected candidate already exists; a later, separately signed confirmatory manifest binds R0/R1 after the development database is locked. Actor and candidate-only verifier traffic require capacity-isolated provider quota pools to prevent inter-arm interference. The primary estimand is the equally task-weighted paired accuracy difference over the fixed 89-task population. A task-blocked analysis aligned to that finite-population estimand, complete invalid-trial accounting, cost per successful trial, a design-stage power check, and a fail-closed non-estimable rule for zero-variance statistics are specified before activation.

\begin{table}[H]
\centering
\caption{Frozen confirmatory designs and activation state. Counts are planned, not observed results.}
\label{tab:confirmatory-protocols}
\scriptsize
\begin{tabular}{p{0.19\textwidth} p{0.28\textwidth} p{0.22\textwidth} p{0.21\textwidth}}
\toprule
\textbf{Protocol} & \textbf{Primary question} & \textbf{Planned primary design} & \textbf{Current state} \\
\midrule
Authority--effect governance & Unauthorized effects and false verification under attacks/faults & 420 sealed scenario clusters; four primary arms; matched benign conditions; three replicate seeds & \hold{}; controlled-runtime and optional provider stages unexecuted \\
Terminal-Bench 2.1 reliability & Task-weighted accuracy and efficiency of one frozen candidate & 89 tasks; five attempts; two interleaved arms; 890 valid trials & \hold{}; no candidate or provider/model binding activated \\
Promotion / production & Independent serving decision and field outcomes & Separate signed promotion and operational protocol required & \hold{}; neither study grants authority \\
\bottomrule
\end{tabular}
\end{table}

The protocol, statistical analysis plan, JSON Schemas, and immutable HOLD manifest are publication artifacts. Their presence improves auditability but does not constitute preregistration. Each development, confirmatory, runtime, or provider stage needs its own independently signed and timestamped ACTIVE manifest binding the artifacts available and required before that stage's first trial; a later manifest cannot retrospectively fill an earlier unknown candidate.

\section{Results}

\subsection{Repository-local conformance}

\begin{figure}[H]
\centering
\includegraphics[width=0.82\textwidth]{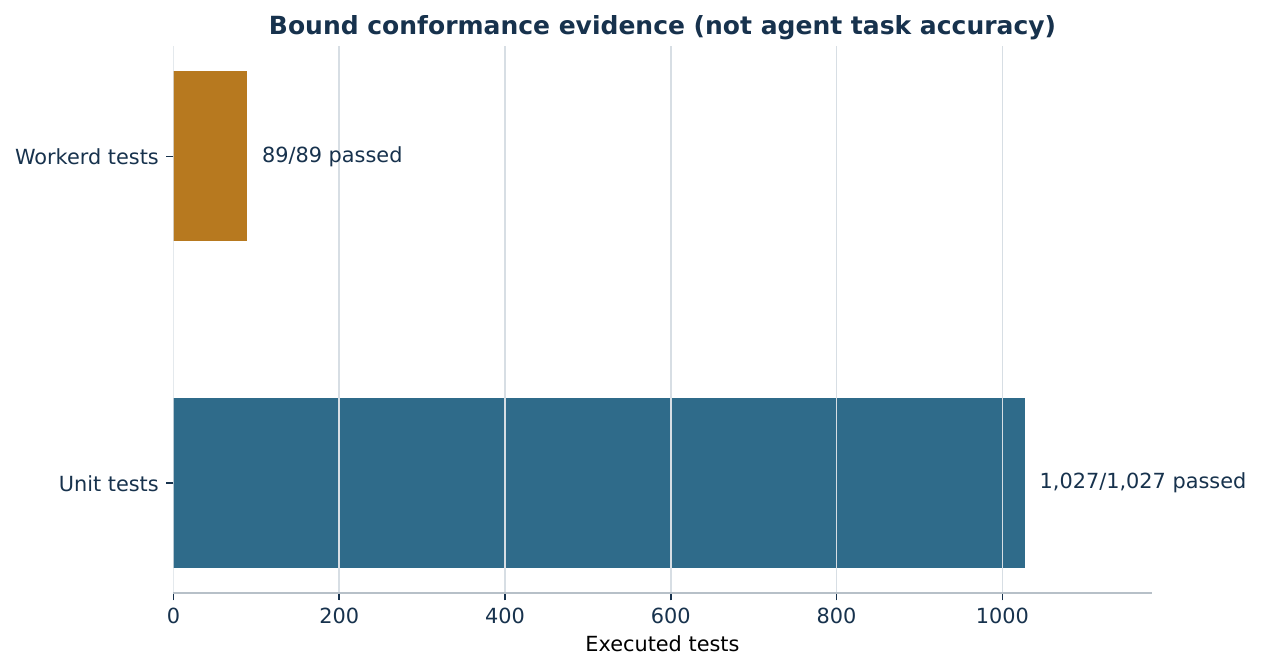}
\caption{Bound conformance evidence. These are executable software tests, not independent agent tasks or production observations.}
\label{fig:test-evidence}
\end{figure}

The author-run release record reports that, at the pinned revision, 1,027 of 1,027 unit tests and 89 of 89 Workerd tests passed. Figure~\ref{fig:test-evidence} visualizes counts only; it must not be read as task accuracy or independent reproduction.

\begin{table}[H]
\centering
\caption{Aggregate coverage at the audited revision. All declared floors passed, and all 363 expected source files appeared in coverage instrumentation.}
\label{tab:coverage}
\small
\begin{tabular}{lrrrrr}
\toprule
\textbf{Measure} & \textbf{Covered} & \textbf{Total} & \textbf{Observed} & \textbf{Floor} & \textbf{Status} \\
\midrule
Statements & 16,051 & 23,573 & 68.09\% & 65\% & \pass \\
Branches & 11,651 & 18,697 & 62.31\% & 50\% & \pass \\
Functions & 2,777 & 3,698 & 75.09\% & 70\% & \pass \\
Lines & 14,972 & 21,133 & 70.84\% & 70\% & \pass \\
\bottomrule
\end{tabular}
\end{table}

\begin{figure}[H]
\centering
\includegraphics[width=0.82\textwidth]{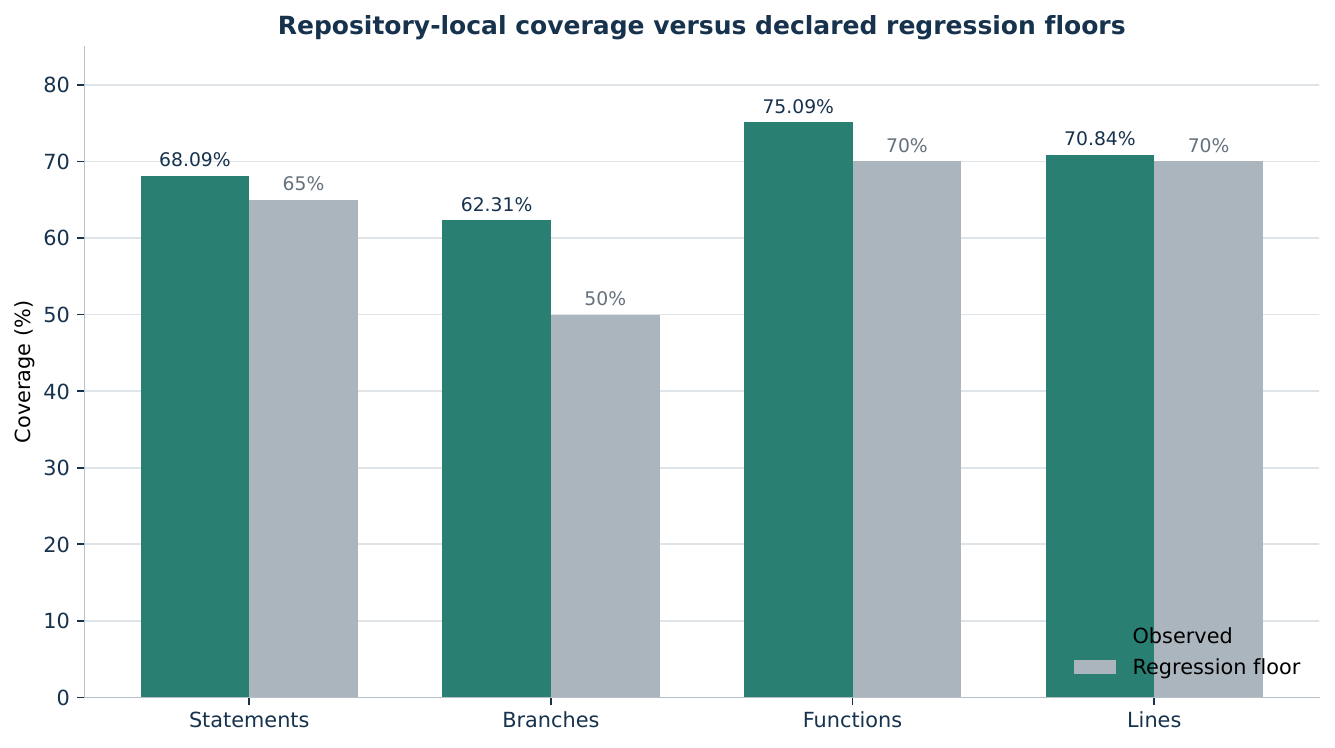}
\caption{Observed repository-local coverage versus declared non-production regression floors. Coverage is a regression signal, not a reliability probability.}
\label{fig:coverage}
\end{figure}

Coverage percentages in the release artifact are the Istanbul reporter's values, truncated to two decimal places; they are not independently rounded from the displayed integer numerators and denominators.

The author-run validation additionally reported checks of 664 source files, 18 canonical action families, 11 aliases, 17 adapter contracts, 41 additive migrations, 28 declared frontier tables, 30 repository topology entries for Workers, 19 for Queues, six explicit HOLD entries, and 527 Markdown links. \repo{npm audit --audit-level=high} found no high- or critical-level finding in the audited dependency state. These counts describe repository declarations; they do not establish that external resources were deployed or healthy.

The release record's production graduation remained \hold{} because no trusted, fresh live promotion receipt was supplied. H1 and H2 are therefore supported only within the implementation and repository-local evidence class.

\subsection{Terminal-Bench pilot}

Both arms passed 17 of 36 strict end-to-end trials, for 47.22\% pass@1. The absolute primary difference was 0.0 percentage points. Recomputed pass@2 was 55.56\% for baseline and 58.33\% for the reliability layer; pass@3 was 58.33\% in both arms. These secondary values describe how the same number of passes was distributed across tasks and do not establish an improvement. Each arm contained 17 passes, 16 scored failures, and three parser-null outcomes. Excluding only the parser-null records yields 17/33 (51.52\%) in each arm. The reliability layer consumed 2,233,805 input and 51,431 output tokens, compared with 1,624,737 input and 34,122 output tokens for baseline. Relative overhead was 37.49\% for input and 50.73\% for output. In the post hoc complete-case diagnostic, input/output totals were 901,131/26,923 for baseline and 1,479,200/44,056 for the reliability layer, increasing overhead to 64.15\%/63.64\%. The excluded task itself consumed 723,606 baseline and 754,605 reliability-layer input tokens, so missingness materially affects resource estimates. An integrity audit also corrected a legacy trace aggregation defect: the rendered summary had retained one trace per task and reported three rollbacks and 13 verification rounds, whereas summing all 36 trial traces yields 13 rollbacks and 42 verification rounds. This correction increases the measured mechanism activity but does not change any task outcome.

\begin{table}[H]
\centering
\caption{Terminal-Bench Core 0.1.1 pilot results. Parser-null records are non-passes in the strict primary outcome; the available-scorer row is sensitivity only.}
\label{tab:pilot}
\small
\begin{tabular}{lrrr}
\toprule
\textbf{Metric} & \textbf{Baseline} & \textbf{Reliability layer} & \textbf{Difference} \\
\midrule
Strict passing trials & 17/36 & 17/36 & 0 \\
Pass@1 (trial accuracy) & 47.22\% & 47.22\% & 0.00 points \\
Pass@2 (task-balanced estimator) & 55.56\% & 58.33\% & +2.78 points \\
Pass@3 (tasks solved at least once) & 58.33\% & 58.33\% & 0.00 points \\
Scored fail / parser-null & 16 / 3 & 16 / 3 & 0 / 0 \\
Available-scorer sensitivity & 17/33 (51.52\%) & 17/33 (51.52\%) & 0.00 points \\
Sensitivity input tokens & 901,131 & 1,479,200 & +64.15\% \\
Sensitivity output tokens & 26,923 & 44,056 & +63.64\% \\
Input tokens & 1,624,737 & 2,233,805 & +37.49\% \\
Output tokens & 34,122 & 51,431 & +50.73\% \\
Agent steps & 384 & 457 & +19.01\% \\
Rollbacks & 0 & 13 & +13 \\
Verification rounds & 0 & 42 & +42 \\
Fallbacks & 0 & 0 & 0 \\
\bottomrule
\end{tabular}
\end{table}

\begin{figure}[H]
\centering
\includegraphics[width=\textwidth]{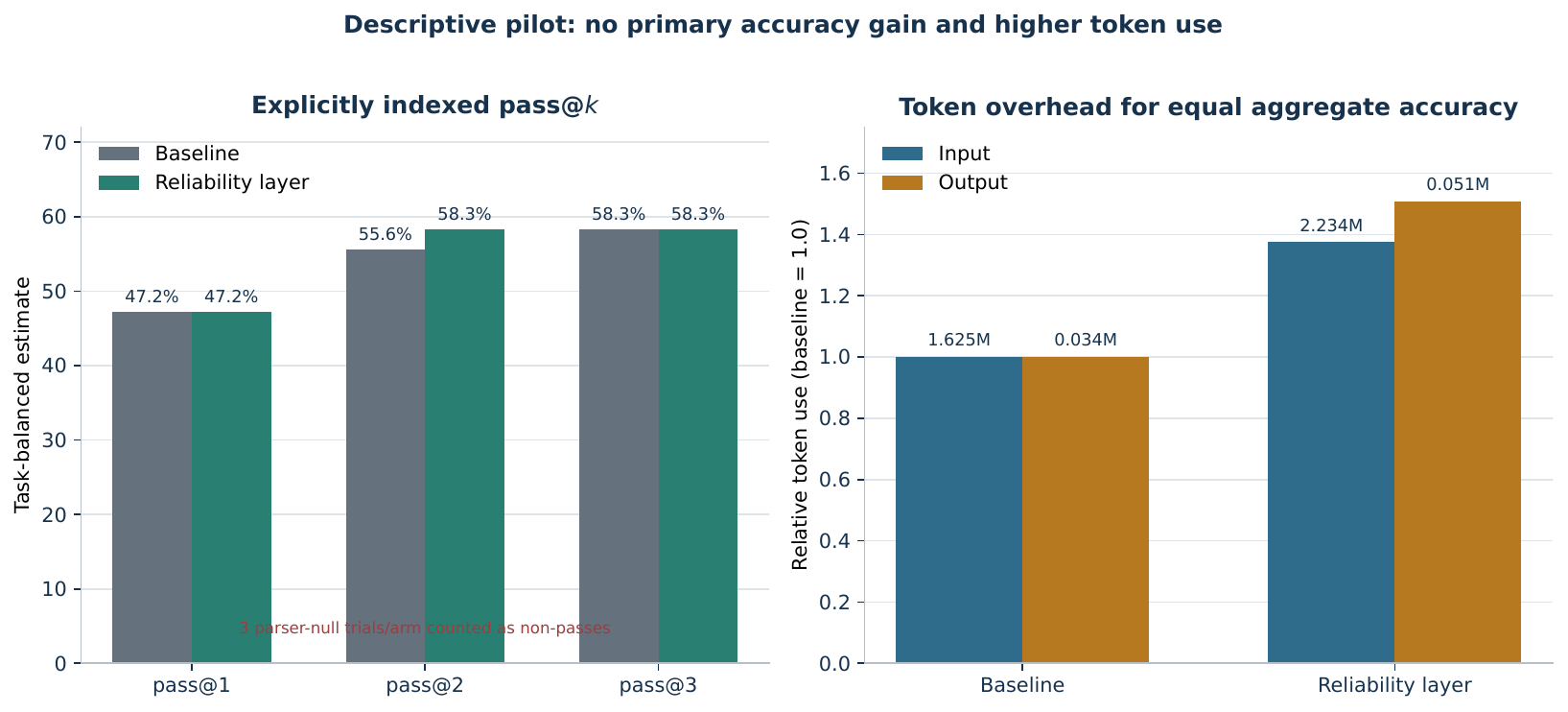}
\caption{Provider-backed pilot. The left panel reports explicitly indexed pass@$k$ estimates; three parser-null records per arm remain non-passes. The right panel normalizes each token type to its baseline while annotating absolute totals.}
\label{fig:pilot}
\end{figure}

\begin{figure}[H]
\centering
\includegraphics[width=0.86\textwidth]{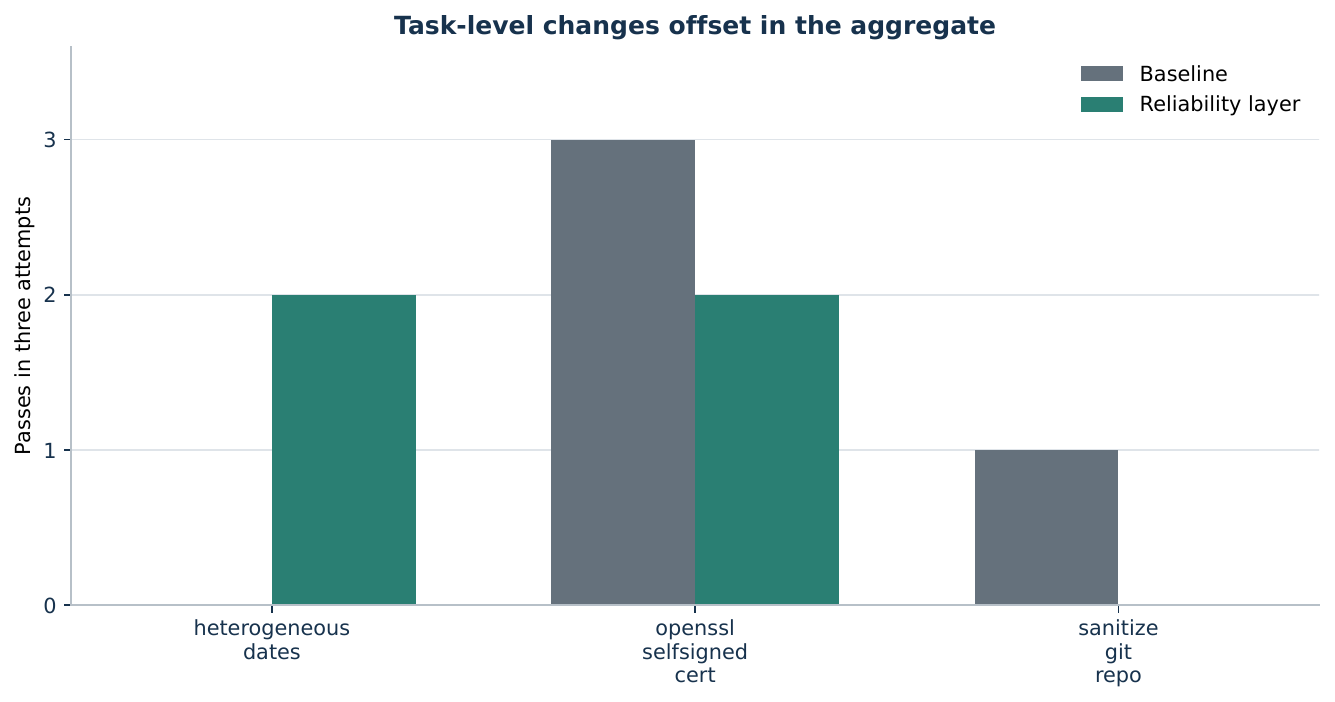}
\caption{The three tasks whose pass counts differed between arms. This diagnostic plot is not evidence that the reliability layer is better for the broader task distribution.}
\label{fig:task-deltas}
\end{figure}

\begin{sloppypar}
Task-level changes were heterogeneous (Figure~\ref{fig:task-deltas}). The reliability layer changed \path{heterogeneous-dates} from 0/3 to 2/3. In the opposite direction, \path{openssl-selfsigned-cert} changed from 3/3 to 2/3 and \path{sanitize-git-repo} from 1/3 to 0/3. The remaining nine tasks were unchanged. Positive and negative changes offset in the aggregate.
\end{sloppypar}

H3 is not supported. Under this model, curated task subset, prompt, and budget, the added reliability mechanisms did not improve the registered descriptive primary outcome and increased token use. The 2.78-point pass@2 difference is a secondary transformation of only 12 task clusters, was not preregistered as primary, and receives no inferential interpretation. The result does not imply that verification or rollback is universally ineffective; it constrains claims about this configuration and motivates separating reliability mechanisms in development ablations.

\subsection{Recursive benchmark gate}

The deterministic conformance artifact records an accepted synthetic candidate: six trial pairs matched, the authored protected and effectiveness metrics were equal, and the expected decision was reproduced while the artifact remained explicitly production-ineligible. This supports H4 only as deterministic conformance.

The later two-order provider-backed coordination-proxy development comparison is reported separately in Table~\ref{tab:recursive-provider}. Both endpoints completed 180/180 trials with mean accuracy 1.000, recovered all 18 hermetically injected crash-after-commit cases, and recorded zero protected violations. Relative to baseline, the source-authored candidate used 37.11\% fewer tokens, 33.84\% lower estimated endpoint cost, and 11.63\% fewer steps. Total trace use across execution and evaluation was 306,330 tokens and an estimated \$2.676842.

\begin{table}[H]
\centering
\caption{Post-debug, two-order provider-backed coordination-proxy development comparison at benchmark revision \texttt{0ec152a4}. Each arm contains 150 deterministic-preflight and 30 model-assisted trials; resource changes are candidate relative to baseline.}
\label{tab:recursive-provider}
\small
\begin{tabular}{p{0.28\textwidth} p{0.20\textwidth} p{0.20\textwidth} p{0.22\textwidth}}
\toprule
\textbf{Metric} & \textbf{Baseline} & \textbf{Candidate} & \textbf{Change} \\
\midrule
Completion & 180/180 & 180/180 & 0 \\
Mean accuracy & 1.000 & 1.000 & 0 \\
Hermetic crash-after-commit recovery & 18/18 & 18/18 & 0 \\
Protected violations & 0 & 0 & 0 \\
p95 latency & 24,561 ms & 15,212 ms & $-9{,}349$ ms ($-38.06\%$), descriptive only \\
Total tokens & 151,315 & 95,157 & $-37.11\%$ \\
Estimated endpoint cost & \$1.578215 & \$1.044142 & $-33.84\%$ \\
Total steps & 258 & 228 & $-11.63\%$ \\
\bottomrule
\end{tabular}
\end{table}

The overall 95\% interval for the p95 difference, computed by a deterministic 2,000-resample fixture-cluster bootstrap over both endpoint orders, was $[-14{,}814,\,+2{,}613]$ ms. Because it crosses zero, the point estimate does not support a latency-improvement claim. The defensible E1 finding is lower token, estimated-cost, and step totals at matched observed conformance in this coordination-proxy development comparison. It is not evidence of improved quality, autonomous recursive optimization, production behavior, or general specialist superiority. The credential-free retained-artifact verifier records \repo{counterbalanceQualification:true}, \repo{productionQualification:false}, \repo{servingPathQualification:false}, and \repo{servingAuthority:none}.

\begin{figure}[H]
\centering
\includegraphics[width=0.96\textwidth]{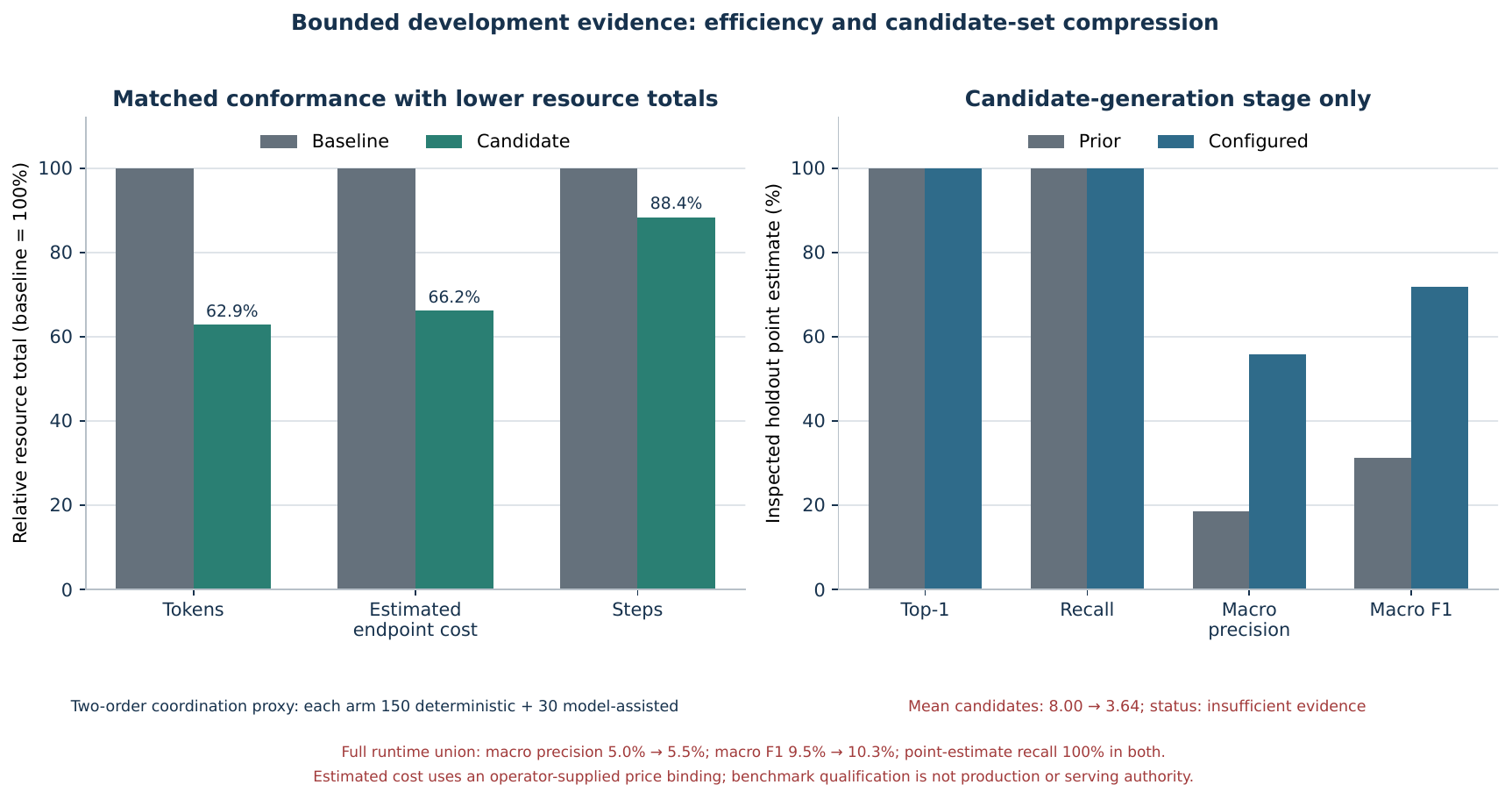}
\caption{Bounded development results. Left: provider-backed coordination-proxy candidate resource totals relative to baseline at matched observed conformance. Right: inspected hybrid candidate-generation point estimates; the downstream runtime-scope union had low macro precision (5.0\% baseline, 5.5\% configured). The latency point estimate is excluded because its interval crosses zero; the selector result is explicitly non-confirmatory.}
\label{fig:development-efficiency}
\end{figure}

\subsection{Learning-loop wiring and hybrid tool selection}

All three worker version probes reported revision \texttt{705abdce}. Checked-in role-specific configuration at that revision enabled reflection, recall accounting, and tool-health feedback on the chat worker and memory compilation on the router and AI workers. This is a source/configuration functionality result only: no invocation or completion was observed, and the audit contains no production estimate of correct learned lessons, memory-retrieval lift, task success, tool success, or user benefit.

On the 44-case inspected development holdout, the bounded candidate generator recorded 100.0\% point-estimate top-1 accuracy and required-tool recall with zero missed requests in both configurations. Candidate-stage macro precision increased from 18.5\% to 55.9\%, macro F1 from 31.2\% to 71.8\%, and the mean candidate count decreased from 8.00 to 3.64 with zero paired complete-case regressions. After the candidates passed through the full deterministic runtime tool-scope union, point-estimate recall remained 100.0\% and misses remained zero in both configurations, but macro precision was only 5.0\% versus 5.5\% and macro F1 9.5\% versus 10.3\%. The latter is a descriptive 0.5-point same-stage precision increase, not high end-to-end precision or task-quality lift.

Across all 114 inspected development cases, the recorded counts were 114/114 top-1 successes, 114/114 complete-case required-tool-recall successes, and 0/114 missed requests. The nominal one-sided 95\% zero-error diagnostic gives a 97.41\% lower bound for top-1 and complete-case recall and a 2.59\% missed-request upper bound. Independence is not established. Reaching the stated 99\% lower-recall and 1\% upper-miss targets would require at least 299 independent zero-error cases globally and separately within every reported family. No multiplicity adjustment was applied across 14 families, so the per-family targets do not constitute a simultaneous family-wise guarantee. The artifact therefore remains \repo{insufficient\_evidence}.

\begin{table}[H]
\centering
\caption{Evidence-lane disposition.}
\label{tab:lane-status}
\small
\begin{tabular}{p{0.25\textwidth} p{0.17\textwidth} p{0.48\textwidth}}
\toprule
\textbf{Lane} & \textbf{Status} & \textbf{Permitted interpretation} \\
\midrule
Pinned release validation & \pass & Repository-local contracts and regression floors passed at the named revision. \\
Terminal-Bench pilot & Completed, observed tie & One small provider-backed comparison found equal accuracy and higher token use; no inferential test was run. \\
Recursive specialist conformance & \pass & Pairing, aggregation, and gate mechanics work on synthetic inputs. \\
Provider coordination proxy & \pass & Producer-recorded engineering benchmark qualification; equal observed conformance and lower token, estimated-cost, and step totals for the source-authored candidate. \\
Learning-loop source/configuration & Observed & Bounded feedback paths are present at the named deployed revision; outcome impact is unmeasured. \\
Hybrid tool selection & Development only & Candidate-stage compression, low full runtime-union precision, and preserved point-estimate recall; statistical qualification remains insufficient. \\
Independent serving promotion & \hold & No serving-authority claim. \\
Production outcomes & \hold & No deployed safety, latency, reliability, self-learning, or user-outcome claim. \\
\bottomrule
\end{tabular}
\end{table}

\section{Discussion}

\subsection{The principal result is a boundary, not a leaderboard number}

Praxa's strongest supported contribution is the representation of a frequently blurred distinction: a model can produce a useful proposal without possessing authority to enact it. The public client describes requests but holds no provider credentials. The gateway derives identity and scope rather than trusting client fields. Policy canonicalizes and admits an action. The broker owns execution authority. Evidence is represented separately from mission projection. Evaluation evidence remains separate from rollout authority.

This does not prove the system safe. It makes narrower questions testable: Can an alias widen the canonical action? Does a resumed mission recheck authorization state? Can an approval card render a different payload than the gate records? Can a synthetic benchmark activate a candidate? Scientific progress is easier when each claim has an explicit oracle.

\subsection{Efficiency without quality separation}

The provider-backed coordination-proxy comparison isolates a useful but narrow systems result. Both endpoints reached the measured conformance ceiling, while the candidate used fewer tokens, lower estimated endpoint cost, and fewer steps. This supports an efficiency interpretation for the named development fixture set; it cannot separate quality because the observed quality measures were equal at their ceiling. The estimated cost uses the run-bound operator-supplied price schedule rather than a provider invoice. The candidate topology was source-authored, so the result is consistent with reduced coordination overhead but does not demonstrate optimizer-generated self-improvement or execution of the production specialist topology.

The deployed learning-loop audit is similarly structural. Feedback paths can record bounded negative-outcome reflection, recall, compiled concepts, and aggregate tool health without widening authority. Whether those paths learn correct lessons or improve future tasks remains an empirical question. The inspected tool-selection result supports a smaller, more precise candidate set at unchanged point-estimate recall. The downstream runtime-scope union, however, had only 5.0\% versus 5.5\% macro precision. This stage separation, the nominal diagnostic bounds, and the tuned labels prevent a high-precision end-to-end or confirmatory interpretation.

\subsection{A descriptive benchmark tie is design evidence}

The Terminal-Bench result contradicts a simple hypothesis that adding planning, self-verification, checkpoint/rollback, and recovery will necessarily improve task accuracy in this pilot. The observed aggregate tie and token overhead suggest at least three mechanisms; they are descriptive evidence, not a statistical equivalence or null-effect finding.

First, verification consumes budget and may redirect a capable trajectory without adding independent information. Second, actor and verifier used the same recorded model binding, so correlated errors can survive self-review; because no immutable provider snapshot was retained, even model-version identity is not independently provable. Third, stochastic trajectory divergence makes a within-arm rollback guarantee different from cross-arm non-inferiority: restoring a reliability-arm checkpoint does not guarantee that its initial trajectory matches the baseline arm's trajectory.

The pilot does show why state-based guards remain valuable. A verification-driven repair can be checkpointed and rolled back if it does not yield a verified improvement within the same trajectory. That invariant limits damage from rework, but it should not be advertised as a statistical guarantee that the reliability arm cannot underperform baseline over a task distribution.

The next capability study should first establish a minimal direct-actor baseline
and sweep accessible frontier models under identical tools, budgets, and
environments; model-driven gains must not be attributed to the harness. Candidate
changes should then be introduced one at a time: structured file operations,
managed background processes, task-aware command timeouts, state-preserving
context compaction, and evidence-triggered verification. Planning, checkpoints,
independent semantic review, and at most one alternate repair trajectory should
activate only after observable triggers such as a failed deterministic check,
tool error, contradictory evidence, explicit uncertainty, stalled progress, or a
high-risk mutation. One global ledger must charge actor, planner, verifier,
compactor, fallback, and repair calls. Development traces may train general
lessons only on a disjoint development partition; a frozen candidate is then
tested on all 89 Terminal-Bench 2.1 tasks at five attempts per arm with task-
blocked inference and anti-reward-hacking trace review. A larger experiment may
find benefit, harm, or continued parity; the present study cannot decide among
them.

\subsection{Why a scalar score is insufficient}

An agent candidate could improve average quality while introducing one authorization failure. A weighted score might allow latency or accuracy gain to numerically offset that failure. Praxa treats authorization, safety, contract, grounding, and structured-output violations as protected dimensions. A candidate must remain non-inferior on effectiveness, remain within resource ceilings, avoid protected regressions, satisfy review requirements, and strictly improve at least one held-out or adversarial raw metric. A reporting score can summarize evidence but has no serving authority.

\subsection{Recursive improvement as governed search}

Recursive harness optimization is best interpreted as bounded candidate generation, not autonomous self-modification. The optimizer may revise role instructions and information flow, but it cannot edit the tools, credentials, policies, approvals, memory boundaries, model binding, or promotion mechanism used to judge it. The controller likewise cannot promote its own result without external authority. This architecture preserves an evaluable separation between optimizer, evaluator, and serving decision even when their implementations share infrastructure. The retained provider comparison evaluated a source-authored candidate and did not invoke the configured optimizer; it is evidence for candidate evaluation and resource accounting, not for successful automated search.

\subsection{Implications for agent publication practice}

We recommend that agent-system papers report a claim--evidence matrix alongside aggregate metrics. Test counts should not appear in the same semantic column as task success. Synthetic fixtures should be labeled in the figure title, caption, and artifact schema. Provider runs should bind exact model identifiers, prompts, budgets, tasks, timestamps, scorers, and raw-result hashes. Promotion and production evidence should be reported independently. When two representations of a metric disagree, the metric should be withheld or reconciled rather than selectively quoted.

\section{Limitations and threats to validity}

\textbf{Construct validity.} Test pass rate and code coverage are proxies for implementation conformance, not direct measures of agent safety or utility. Coverage records executed syntax and branches, not assertion adequacy.

\textbf{Internal validity.} The release audit was executed by the project environment rather than an independent laboratory. Cryptographic hashes identify artifacts but do not certify correct test design. The Terminal-Bench arms shared a model and base loop, yet provider stochasticity and trajectory divergence remain confounders. No blinding protocol was applied to engineering iteration before the final run. The pilot used a fixed, non-interleaved arm order, so provider drift and order effects are not separable from treatment. The baseline permitted rebuild while the later reliability arm used \repo{no\_rebuild:true}; build/cache state was therefore not held symmetric. Verifier calls were not fully charged to the soft loop budget, creating treatment-specific resource under-accounting despite equal nominal configuration. The archive records temperature 0.0 and within-arm concurrency three, but not top-$p$, request seeds, provider-side defaults, host/runtime versions, or container-image digests.

\textbf{Outcome integrity.} Six pilot records have no parser result. Counting them as non-passes is conservative for completion but cannot reveal whether the task implementation, scorer, or agent caused the missing value. The available-scorer sensitivity is potentially selection-biased and is not the primary result. The earlier unindexed \repo{pass\_at\_k} label also demonstrates why a metric name without $k$ is not reproducible even when the underlying counts are recoverable.

\textbf{Statistical validity.} The Terminal-Bench provider pilot used 12 curated tasks and three attempts per task. Attempts are clustered by task, and no confirmatory sample-size calculation was preregistered. We therefore report descriptive values and no confidence interval or significance test. The coordination-proxy comparison used 30 fixed fixtures with three repetitions in each of two endpoint orders; 150 of 180 trials per endpoint terminated deterministically and both arms reached the observed quality ceiling. Counterbalancing reduces order confounding but does not create a new task population. Resource totals therefore support a bounded development comparison, not a population-level quality or latency estimate. The tool-selection zero-error bounds are nominal diagnostics: case independence is not established, and no multiplicity correction provides simultaneous coverage across the 14 families. The frozen Terminal-Bench 2.1 plan improves the external design but has produced no data.

\textbf{External validity.} The Terminal-Bench pilot concerns one recorded gateway/provider alias, one harness configuration, and a curated subset of the legacy Core 0.1.1 suite; it is not directly comparable with the official 89-task Terminal-Bench 2.1 leaderboard. The later coordination-proxy comparison records an execution alias, a configured but unused optimizer binding, three evaluator identifiers, and the resolved identifiers observed in both order artifacts. \repo{gpt-5.6-sol} remains a requested alias rather than an archived immutable provider snapshot. Results need not generalize to other models, providers, topologies, environments, risk levels, or user populations.

\textbf{Synthetic and provider conformance.} The original recursive artifact's authored inputs prove arithmetic, replay, matching, and gating behavior only. The later provider result is a separate post-debug coordination-proxy development comparison. Earlier development executions informed engineering changes that exposed the observed deterministic policy disposition to model-visible context and required exact reproduction. The successful partition was therefore not an untouched holdout. The task partition was fixed and visible, evaluator judgments came from models rather than independent human review, and the candidate was source-authored. The two endpoint orders reduce a known order confound but do not remove provider drift or development-set adaptation. The candidate-first workflow's original red conclusion and later credential-free repair are both retained. Recovery was a hermetic crash-after-commit injection; the production \repo{spawn\_agents} topology and web/mobile serving paths were not executed. The result supports lower resource totals at matched observed conformance, not improved quality, latency, autonomous recursive optimization, production performance, or generalization.

\textbf{Learning-loop evidence.} Source and deployment-revision parity show role-specific configuration for bounded reflection, recall accounting, memory compilation, and tool-health feedback paths; they do not establish that every path was enabled on every worker or that an invocation completed. They do not show that a reflection was correct or that later retrieval, tool choice, task success, or user outcomes improved. Recall accounting is started without awaiting completion, the tool-health snapshot is cached per isolate, and no retained real-corpus memory-retrieval or production outcome A/B was found. The tool-selection labels and cases were inspected during tuning; its nominal one-sided bounds remain below the release targets, and its full runtime-union macro precision is low.

\textbf{Deployment evidence.} Repository topology is not a deployment inventory. This paper reports no live availability, queue-delivery rate, production throughput, active-user count, incident reduction, or user satisfaction.

\textbf{Benchmark contamination.} The specialist task partition is not a secret, unseen population. Public-benchmark contamination cannot be excluded without provider training-data evidence or sealed dynamic holdouts.

\textbf{Evaluator dependence.} Same-model self-verification can reproduce actor errors. Distinct evaluator configurations and randomized orientation can mitigate but not eliminate model-family correlation, verbosity preference, or rubric leakage.

\textbf{Private reproducibility boundary.} Security-sensitive harness code and raw provider traces remain access-controlled. The public can inspect the paper, redacted aggregates, hashes, figures, protocols, public SDK, build pipeline, and invariant--test map, but cannot independently rerun every private authority path or reconstruct every provider trajectory without authorized access. The provider-development aggregate publishes run chronology and retained-file hashes, but the first two development executions retained no result archive and the private retained artifact is time-limited. The repository-local conformance summaries still omit raw Vitest/Workerd reports, command transcripts, per-test transcripts, durations, and a trusted independent signature. This is a material limitation, not a complete open-reproducibility claim.

\textbf{Prospective-protocol boundary.} Freezing endpoints, invalidity rules, randomization, and analysis reduces future researcher degrees of freedom; it does not validate the proposed simulator, establish sufficient power, supply an immutable provider model, confer independent custody, or turn a HOLD manifest into a preregistration. The released relational checker also does not yet establish cross-artifact chronology and signature trust, resolve provider receipts, enforce redaction provenance, recompute governance summaries from raw ledgers, or retrieve and hash every referenced artifact. Those conditions are explicit activation blockers, not completed validations.

\textbf{Secondary utilization audit.} The multi-runtime application's source tree was dirty and changing during audit. Its deterministic counts are included to map utilization and claim ceilings, not as primary publishable evidence. A confirmatory publication should rerun them from a clean, pinned revision and archive the generated artifacts.

\textbf{No formal verification or human-subject study.} Preliminary specifications have no reported model-checking output. No usability, accessibility, fairness, reliance, or field-impact study was conducted.

\textbf{Temporal validity.} Models, dependencies, prices, standards, and hosted systems change. Confirmatory evaluations should bind immutable source and model revisions; moving aliases are inadequate for durable claims.

\section{Ethics, safety, and responsible disclosure}

Consequential agent actions can affect privacy, finances, communications, employment, health-related workflows, and organizational records. Praxa therefore treats consent, approval, purpose, tenant identity, resource budget, target, and revocation as server-owned constraints. The design also preserves uncertainty rather than presenting incomplete external evidence as success.

Governance mechanisms can create false assurance when engineering evidence is overstated. This paper publishes negative status beside positive results: the Terminal-Bench pilot did not improve aggregate accuracy; synthetic values are labeled synthetic; repository counts are not deployments; formal verification is not claimed; and the final coordination-proxy development run satisfied a producer-recorded engineering benchmark gate while production qualification, serving authority, and field outcomes remained absent.

No customer data, user conversation, or human-subject judgment was required for the reported aggregates. Raw transcripts are excluded because they can contain benchmark prompts, model completions, credentials, or operational details. Future trace releases should undergo explicit license, privacy, authority, and redaction review.

\section{Reproducibility and artifact availability}

The canonical LaTeX source, bibliography, flat-color figures, machine-readable aggregates, frozen protocols, JSON Schemas, build scripts, PDF, DOCX, and arXiv source archive are published at:

\begin{center}
\href{https://github.com/praxa-labs/praxa-benchmarks}{https://github.com/praxa-labs/praxa-benchmarks}
\end{center}

The immutable publication target for this revision is \href{https://github.com/praxa-labs/praxa-benchmarks/releases/tag/preprint-v1.4.0}{preprint-v1.4.0}; release assets are checksummed in \repo{output/publication-manifest.json}. Citations should use the tag or release rather than a mutable \repo{tree/main} path.

The repository-local result is bound to the following immutable identifiers:

\begin{itemize}
  \item source revision: \small\nolinkurl{69cacd7f5c01cec311aec05efa8e66edc7a117ca}\normalsize;
  \item evidence bundle: \small\nolinkurl{b204c62cc2b77f1027a01e2efd8c36ba1883a21f199379f81f654ce6ac1a676f}\normalsize.
  \item invariant/source/test/oracle map: \nolinkurl{paper/data/invariant-test-map.json}.
\end{itemize}

The Terminal-Bench public aggregate includes the exact study revision, aggregation revision, run interval, task/attempt counts, model binding, budget, outcome taxonomy, recomputed pass@1/2/3, and raw-result hashes. The provider-history aggregate binds the earlier coordination-proxy development chronology and gate contract. The counterbalanced aggregate adds both endpoint orders, workflow conclusions, retained-artifact identifiers and hashes, repaired credential-free verification, resolved model identifiers, resource deltas, and qualification state. The learning-loop artifact binds deployed source/configuration evidence and the separately staged candidate-generation and runtime-union point estimates with explicit \repo{confirmatory:false} status. The build validator recomputes the central quantities, rejects a latency claim when its interval crosses zero, checks that benchmark qualification remains separate from serving authority, validates the protocol status contracts, and recompiles the exact arXiv archive.

Re-running the full private conformance suite needs authorized access to the pinned implementation. Re-running the provider pilot additionally requires the benchmark containers and model provider. Rebuilding the publication requires no private credential.

No reuse license is attached to the repository. Public availability permits review but does not itself grant reuse or redistribution rights to the code or data. The author selected CC BY 4.0 for the arXiv article distribution; that selection does not license the repository artifacts.

\section{Conclusion}

Praxa demonstrates that governance claims for AI agents can be translated into executable contracts and evidence artifacts spanning proposal, authority, dispatch, external effect, reconciliation, and promotion. For the audited snapshot, the author-run record reports all 1,027 unit tests and 89 Workerd tests passing, every one of 363 expected source files appearing in coverage instrumentation, and all four declared coverage floors met; the public invariant map exposes the intended claim-to-oracle links while the absence of raw run transcripts limits independent audit. The separate Terminal-Bench pilot produced a bounded descriptive tie: the reliability layer matched baseline at 17/36 strict passing trials while using 37.49\% more input and 50.73\% more output tokens. Explicit pass@$k$ labels and parser-null sensitivity repair an ambiguity in the first artifact without changing the primary conclusion.

The post-debug, two-order coordination-proxy development comparison reached equal observed conformance in both arms while the source-authored candidate used 37.11\% fewer tokens, 33.84\% lower estimated endpoint cost, and 11.63\% fewer steps. The p95 point estimate fell 38.06\%, but its interval crossed zero. The candidate-first execution retained valid evidence but its original workflow ended red at a comparison defect; a later credential-free verifier repaired the comparison and qualified the checksum-bound aggregate. The configured optimizer was not invoked, the production specialist topology was not executed, and neither production nor serving qualification passed. Deployed reflection, recall-accounting, memory-compilation, and tool-health paths establish bounded learning-loop functionality. The inspected hybrid selector provides development-only candidate-stage compression evidence, but its full runtime-union macro precision was 5.0\% versus 5.5\%; neither result establishes production self-learning or task-quality benefit.

The evidence therefore supports implementation and conformance claims, not adversarial security, production safety, universal reliability, or model superiority. The released confirmatory protocols specify the next tests, but only executed, independently reviewable evidence can answer them. The central methodological result is that model proposals, authorization decisions, external effects, evaluation evidence, and serving authority should be linked by attributable transitions and never treated as interchangeable.

\section*{Acknowledgments}

The author thanks the maintainers of the open-source agent, evaluation, security, provenance, and standards artifacts cited in this work. No external funding source is reported for this preprint.

\bibliographystyle{plainnat}
\bibliography{references}

\appendix
\section{Claim--evidence matrix}

\small
\begin{longtable}{p{0.26\textwidth} p{0.27\textwidth} p{0.37\textwidth}}
\caption{Public claim boundary.}\label{tab:claim-matrix}\\
\toprule
\textbf{Claim} & \textbf{Supporting evidence} & \textbf{Excluded interpretation} \\
\midrule
\endfirsthead
\toprule
\textbf{Claim} & \textbf{Supporting evidence} & \textbf{Excluded interpretation} \\
\midrule
\endhead
Praxa exposes public SDK, CLI, MCP contract, and OpenAPI surfaces. & Public \repo{praxa-labs/praxa} sources and package tests. & Does not expose private credentials or prove hosted availability. \\
Model and client outputs do not directly authorize effects. & Gateway/policy/broker contracts and tests. & Not proof against compromise of trusted policy, broker, or signing components. \\
The author-run release contract reported pass. & Named commit, evidence bundle, 1,027/1,027 unit and 89/89 Workerd tests, plus invariant/test mapping. & Raw per-test transcripts and independent reproduction are unavailable; not a timeless guarantee. \\
Coverage inventory was complete. & 363/363 expected files, no missing source files. & Not 100\% statement, branch, function, or line coverage. \\
The provider pilot was executed. & 72 timestamped trials across two arms and raw-result hashes. & Does not validate production Praxa or support aggregate superiority. \\
The reliability layer improved aggregate pilot accuracy. & No supporting evidence; 17/36 in both arms. & Claim rejected for this configuration. \\
Pilot pass@$k$ is reproducibly indexed. & Task-level counts recompute pass@1, pass@2, and pass@3; parser-null outcomes are separately counted. & Does not repair missing raw scorer output or justify inference. \\
The recursive conformance gate accepted its synthetic candidate. & Deterministic paired artifact and expected gate decision. & No real quality, latency, token, cost, or recovery gain. \\
The retained coordination-proxy candidate used fewer resources at matched observed conformance. & Two-order provider-backed post-debug comparison: 180/180 completed trials per arm, mean accuracy 1.000, hermetic recovery 18/18, zero protected violations, and reductions of 37.11\% in tokens, 33.84\% in estimated endpoint cost, and 11.63\% in steps. & Development proxy only; production specialist and web/mobile serving paths were not executed; no quality, latency, production, user-benefit, or general-specialist superiority claim. \\
The retained coordination-proxy candidate satisfied the recorded counterbalance gate. & Credential-free retained-artifact verification over opposite-order runs at benchmark revision \texttt{0ec152a4}; public workflow outcomes, hashes, and aggregate recomputation. & One source workflow ended red after evidence retention; raw private trials cannot be publicly re-adjudicated; production and serving-path qualification remained false and serving authority remained \repo{none}. \\
Bounded learning-loop paths were configured at the observed deployed revision. & All three workers reported revision \texttt{705abdce}; role-specific configs enabled reflection/recall/tool health on chat and memory compilation on router/AI. & Version/configuration evidence only; no invocation, completion, measured self-learning, recall, task, tool-success, or user-benefit lift. \\
The configured hybrid candidate generator compressed its inspected candidate set. & On the 44-case inspected holdout, point-estimate required-tool recall remained 100\%, candidate-stage macro precision rose from 18.5\% to 55.9\%, macro F1 from 31.2\% to 71.8\%, and mean candidates fell from 8.00 to 3.64. & Full runtime-union macro precision was 5.0\% versus 5.5\%; non-confirmatory and \repo{insufficient\_evidence}, not high end-to-end precision, task success, or reranker quality. \\
The frozen protocols are completed experiments. & No confirmatory trial exists; preregistration manifest remains HOLD. & Protocol availability is not empirical evidence. \\
Praxa improves production safety or user outcomes. & No controlled deployment or human-subject study. & \notest{}. \\
Praxa is formally verified. & No completed model-checking artifact. & \notest{}. \\
\bottomrule
\end{longtable}
\normalsize

\section{Terminal-Bench task-level results}

\begin{table}[H]
\centering
\caption{Passes in three attempts for each pilot task. The three \texttt{cron-broken-network} records in each arm were parser-null and count as non-passes. Task identifiers are inherited from Terminal-Bench Core 0.1.1.}
\small
\begin{tabular}{lrrr}
\toprule
\textbf{Task} & \textbf{Baseline} & \textbf{Reliability layer} & \textbf{Delta} \\
\midrule
fix-permissions & 3/3 & 3/3 & 0 \\
csv-to-parquet & 3/3 & 3/3 & 0 \\
heterogeneous-dates & 0/3 & 2/3 & +2 \\
openssl-selfsigned-cert & 3/3 & 2/3 & $-1$ \\
sqlite-db-truncate & 0/3 & 0/3 & 0 \\
decommissioning-service-with-sensitive-data & 0/3 & 0/3 & 0 \\
new-encrypt-command & 2/3 & 2/3 & 0 \\
nginx-request-logging & 0/3 & 0/3 & 0 \\
processing-pipeline & 2/3 & 2/3 & 0 \\
cron-broken-network & 0/3 & 0/3 & 0 \\
sanitize-git-repo & 1/3 & 0/3 & $-1$ \\
fix-pandas-version & 3/3 & 3/3 & 0 \\
\midrule
\textbf{Total} & \textbf{17/36} & \textbf{17/36} & \textbf{0} \\
\bottomrule
\end{tabular}
\end{table}

\section{Public artifact manifest}

\begin{table}[H]
\centering
\caption{Publication artifacts and their role.}
\scriptsize
\begin{tabular}{p{0.41\textwidth} p{0.49\textwidth}}
\toprule
\textbf{Path} & \textbf{Purpose} \\
\midrule
\nolinkurl{paper/main.tex} & Canonical manuscript source. \\
\nolinkurl{paper/references.bib} & Primary-source bibliography. \\
\nolinkurl{paper/data/validation-summary.json} & Redacted pinned conformance aggregate. \\
\nolinkurl{paper/data/invariant-test-map.json} & Pinned invariant, source, test, oracle, and public-evidence limitation map. \\
\nolinkurl{paper/data/terminal-bench-pilot.json} & Redacted provider-pilot aggregate and raw-result hashes. \\
\nolinkurl{paper/data/pilot-integrity-audit.json} & Read-only parser-null, pass@$k$, trace-counter, and source-hash reconciliation. \\
\nolinkurl{paper/data/recursive-harness-conformance.json} & Earlier deterministic synthetic-conformance mechanics; explicitly superseded as a provider-result artifact. \\
\nolinkurl{paper/data/provider-specialist-run-history.json} & Redacted provider-backed coordination-proxy sequence, retained efficiency comparison, gate contract, pricing inputs, and private-artifact hashes. \\
\nolinkurl{paper/data/provider-specialist-}\linebreak[4]\nolinkurl{counterbalanced-confirmation.json} & Two-order workflow outcomes, checksum provenance, credential-free aggregate verification, resource deltas, and qualification boundaries. \\
\nolinkurl{paper/data/learning-loop-tool-selection.json} & Deployed learning-loop source/configuration evidence and non-confirmatory staged hybrid tool-selection evaluation. \\
\nolinkurl{paper/data/utilization-evidence.json} & Secondary multi-runtime application inventory and explicit dirty-tree caveat. \\
\nolinkurl{paper/protocols/} & Frozen prospective governance and Terminal-Bench 2.1 protocols; current state HOLD. \\
\nolinkurl{paper/schemas/} & Machine-readable preregistration, trial-record, database-lock, and content-bound reconciliation contracts. \\
\nolinkurl{paper/PIPELINE-IMPROVEMENT-ROADMAP.md} & Prospective capability-engineering hypotheses, rejection criteria, and evaluation gates; no new result. \\
\nolinkurl{paper/scripts/build_figures.py} & Deterministic figure generation. \\
\nolinkurl{paper/scripts/build_publication.py} & PDF, DOCX, and arXiv archive production. \\
\nolinkurl{paper/scripts/validate_publication.py} & Claim, artifact, PDF, DOCX, and archive checks. \\
\nolinkurl{output/pdf/praxa-harness-preprint.pdf} & Submission-readable PDF. \\
\nolinkurl{output/doc/praxa-harness-preprint.docx} & Editable manuscript rendering. \\
\nolinkurl{output/arxiv/praxa-harness-arxiv-source.zip} & Deterministic arXiv source package. \\
\nolinkurl{output/publication-manifest.json} & Publication metadata plus source and artifact checksums. \\
\bottomrule
\end{tabular}
\end{table}

\end{document}